\documentclass[runningheads]{llncs}

\usepackage{eccv}

\usepackage{eccvabbrv}

\usepackage[table]{xcolor}
\usepackage{graphicx}
\usepackage{booktabs}

\usepackage[accsupp]{axessibility}  %

\usepackage{hyperref}

\definecolor{olivegreen}{rgb}{0, 0.6, 0}
\definecolor{aquamarine}{rgb}{0.05, 0.6, 0.72}

\usepackage{array, makecell}
\usepackage{nicefrac}       %

\usepackage{comment}
\usepackage{xspace}         %
\usepackage{graphicx}%
\usepackage{lipsum}
\usepackage{multirow}
\usepackage{enumitem}
\usepackage{algorithm}
\usepackage{algorithmic}
\usepackage{booktabs}
\usepackage{xcolor}
\usepackage{graphicx}     %
\usepackage{subcaption}   %
\usepackage{caption}      %
\usepackage{amsmath}
\usepackage{amssymb}
\usepackage{wrapfig}
\usepackage[capitalize,noabbrev]{cleveref}
\crefname{section}{section}{sections}
\Crefname{section}{Section}{Sections}
\crefname{subsection}{subsection}{subsections}
\Crefname{subsection}{Subsection}{Subsections}

\usepackage{pifont}%
\newcommand{\cmark}{\color{olivegreen}\ding{51}}%
\newcommand{\xmark}{\color{red}\ding{55}}%

\usepackage{tikz}

\newcommand{\JL}[1]{{\color{olivegreen}[\textbf{\sc JLee}: \textit{#1}]}}
\newcommand{\KH}[1]{{\color{purple}[\textbf{\sc KH}: \textit{#1}]}}
\newcommand{\HY}[1]{{\color{blue}[\textbf{\sc HY}: \textit{#1}]}}
\newcommand{\SJ}[1]{{\color{orange}[\textbf{\sc SJ}: \textit{#1}]}}
\newcommand{\DI}[1]{{\color{aquamarine}[\textbf{\sc DI}: \textit{#1}]}}
\newcommand{\SG}[1]{{\color{red}[\textbf{\sc SG}: \textit{#1}]}}

\def\final{}   %
\ifdefined\final
\renewcommand{\JL}[1]{}
\renewcommand{\KH}[1]{}
\renewcommand{\HY}[1]{}
\renewcommand{\SJ}[1]{}
\renewcommand{\DI}[1]{}
\renewcommand{\SG}[1]{}

\fi

\usepackage{siunitx}

\newcommand{\name}{STARLINC\xspace}
\newcommand{\gradcam}{heatmap\xspace}
\newcommand{\gradcams}{heatmaps\xspace}

\usepackage{orcidlink}

\begin{document}

\title{STARLINC: Satellite Trail Artifact Removal using Inter-Frame Correlation} 

\titlerunning{STARLINC}

\author{Shingeon Kim\inst{1}\orcidlink{0009-0002-1079-8178} \and
Hyeyoon Lee\inst{1}\orcidlink{0000-0003-3130-7921} \and
Dain Kwon\inst{1}\orcidlink{0009-0003-4822-5823} \and
Kanghyun Choi\inst{1}\orcidlink{0009-0005-1056-1316} \and
Sunjong Park\inst{1}\orcidlink{0009-0004-4357-3316} \and
Mi-Ryang Kim\inst{2}\orcidlink{0000-0002-1408-7747} \and
Jeong-Eun Lee\inst{2}\orcidlink{0000-0003-3119-2087} \and
Jinho Lee\inst{1}\orcidlink{0000-0003-4010-6611}}

\authorrunning{S. Kim et al.}

\institute{Department of Electrical and Computer Engineering, Seoul National University 
\and
Department of Physics and Astronomy, Seoul National University \\
\email{\{happyriosshs, hylee817, dain.kwon, kanghyun.choi, ryan0507, mi.ryang, lee.jeongeun, leejinho\}@snu.ac.kr}
}

\maketitle

\begin{abstract}

    The rapid expansion of low Earth orbit satellites such as Starlink is increasingly contaminating astronomical surveys.
    In practice, contaminated images are often identified through inspection. 
    However, modern surveys generate terabytes of data each night, making manual screening infeasible and necessitating reliable automated methods for satellite trail removal.
    Unfortunately, existing general-domain line detection methods fail to generalize to astronomical images due to domain mismatch, which are mostly grayscale with sparse bright stars and have a low signal-to-noise ratio.  
    Moreover, training new models from scratch is impractical due to the lack of large-scale annotated astronomical datasets.
  To address these challenges, we introduce \name, the first ML-based framework for satellite trail removal without requiring tedious pixel-level annotation of astronomical images.   
\name combines synthetic satellite trail generation for training, inter-frame differential maps from temporally adjacent exposures to highlight transient trails, and \gradcams to provide additional localization cues for pixel-level segmentation.
Extensive experiments on real-world data demonstrate substantial improvements over baselines, establishing \name as a scalable solution for next-generation astronomical surveys. 
Code is available at \url{https://github.com/starioKim/STARLINC}.

\end{abstract}

\section{Introduction}
\label{sec:intro}
\begin{wrapfigure}{r}{0.3\textwidth}
  \centering
  \vspace{-16mm}
  \includegraphics[width=\linewidth]{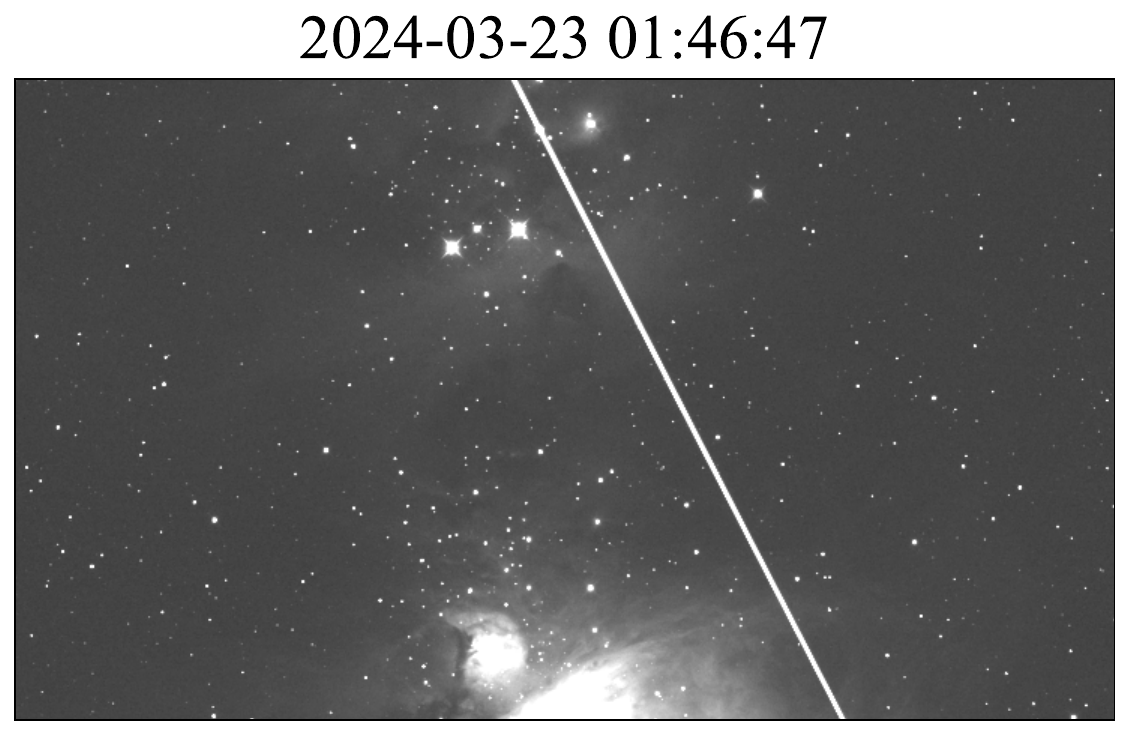}
  \vspace{-7mm}
  \caption{An example of contaminated images.}
  \label{fig:intro}
  \vspace{-9mm} %
\end{wrapfigure}
Satellite contamination~\cite{mroz2022impact} has emerged as a significant and rapidly worsening obstacle to astronomical surveys\footnote{Throughout this paper, we use \textit{astronomical survey} to denote datasets of telescopic images acquired for scientific analysis.}. 
This challenge is driven primarily by the unprecedented growth of low Earth orbit (LEO) satellite constellations~\cite{mcdowell2020low,hainaut2020impact}. 
With the number of active LEO satellites now exceeding 12,000 and increasing at an annual rate of approximately 30\%~\cite{bassa2022analytical,starlink}, satellite trails are no longer rare artifacts but a recurring source of image degradation across modern surveys. 
Previous studies estimate that they affect about 18\% of astronomical survey images~\cite{mroz2022impact,starlink}, and recent SPHEREx observations report contamination in \(73.3^{+1.3}_{-1.2}\%\) of images, with \(2.18^{+0.11}_{-0.09}\) trails per exposure on average~\cite{borlaff2026spherex}.
These contaminations occur when satellites reflect sunlight during exposure, producing bright linear trail segments (see \cref{fig:intro}). %
Such trails introduce artificial light flux into the observations, leading to substantial errors in downstream astrophysical measurements~\cite{borlaff2025satellite,hainaut2020impact}.

Unfortunately, the typical astronomical research pipeline still relies on manual inspection, due to the lack of a reliable automated trail removal method. 
As modern astronomical surveys generate terabytes of data each night, manual inspection can no longer keep pace with the volume of incoming images. 
Moreover, once an image is identified as containing a satellite trail, manual workflows typically exclude the entire image from the dataset, resulting in substantial loss of otherwise valuable observational data.

One might expect satellite trails would be easy to recognize for modern computer vision methods, as the trails appear as simple straight lines against a starry, mostly black sky.
Contrary to this intuition, our empirical evaluation shows that existing line detection methods remain largely ineffective on real astronomical data.
Classical approaches (Hough transform~\cite{hough,canny}, Radon-based detectors~\cite{radon}) are highly sensitive to background noise. 
Several deep learning-based general domain line detectors~\cite{deeplsd,deephough,deepline} fail to reliably detect satellite trails, as they do not account for the extreme sparsity and distinct photometric statistics of astronomical imagery. %
Meanwhile, astronomy-specific approaches~\cite{asta,streakML,unetLSD,yolosatrail} typically rely on pixel-level annotations and exhibit limited generalization. 
Despite strong performance on their original benchmarks, these methods degrade substantially when evaluated on unseen datasets.

To this end, we propose \name, the first ML-based framework for satellite trail removal without requiring pixel-level annotation of astronomical images. 
\name performs pixel-level segmentation to selectively mask contaminated pixels while preserving valid astronomical data.

Through empirical analyses of datasets, we identify two key characteristics:
\begin{enumerate}
    \item The satellite trails typically appear as sharp, narrow lines that occupy only a small fraction of the image.
    \item Astronomical surveys acquire multiple frames as a time series for noise reduction, which are largely identical.
\end{enumerate}
For conventional machine learning methods, both characteristics can be detrimental: the high similarity among frames may confuse the model during training, while the weak signals of thin trails are easily obscured by other image structures.

In contrast, \name is explicitly designed to exploit these properties, resulting in a simple yet effective framework.
First, we provide a synthetic satellite trail generation method to enable effective training of a trail segmentation model without tedious manual pixel-level annotation. 
Secondly, \name leverages temporal inter-frame differential maps across consecutive observations to focus on the transient satellite trail. 
Lastly, we obtain \gradcams highlighting trail-relevant regions from a classifier trained to predict trail presence, and use them as additional visual cues for pixel-level segmentation.
We conduct extensive experiments on real astronomical datasets from an ongoing astronomical survey~\cite{7dt, graham2019zwicky}, where \name consistently outperforms existing baselines by up to +102\% relative mIoU improvement over satellite trail segmentation.
We further evaluate cross-dataset transfer to ZTF~\cite{graham2019zwicky} and NGC-region~\cite{7dt} observations, showing that a \name generalizes to unseen observing conditions.

  \noindent{}Our contributions are summarized as follows:
  \begin{itemize}
  \item We propose \name, a practical satellite trail removal framework that preserves scientifically valid content by masking only contaminated pixels. %
  \item We reduce reliance on pixel-level annotations by introducing synthetic trail generation for training and enhancing localization via temporal inter-frame differential maps and pseudo-localization via \gradcams.
  \item We validate \name on real 7DT survey data~\cite{7dt}, demonstrating up to $+102\%$ relative mIoU over the strongest baseline, and further show cross-dataset transfer to ZTF~\cite{graham2019zwicky} and NGC-region~\cite{7dt} observations.  

  \end{itemize}

\begin{figure}[t]
\centering
\includegraphics[width=\textwidth]{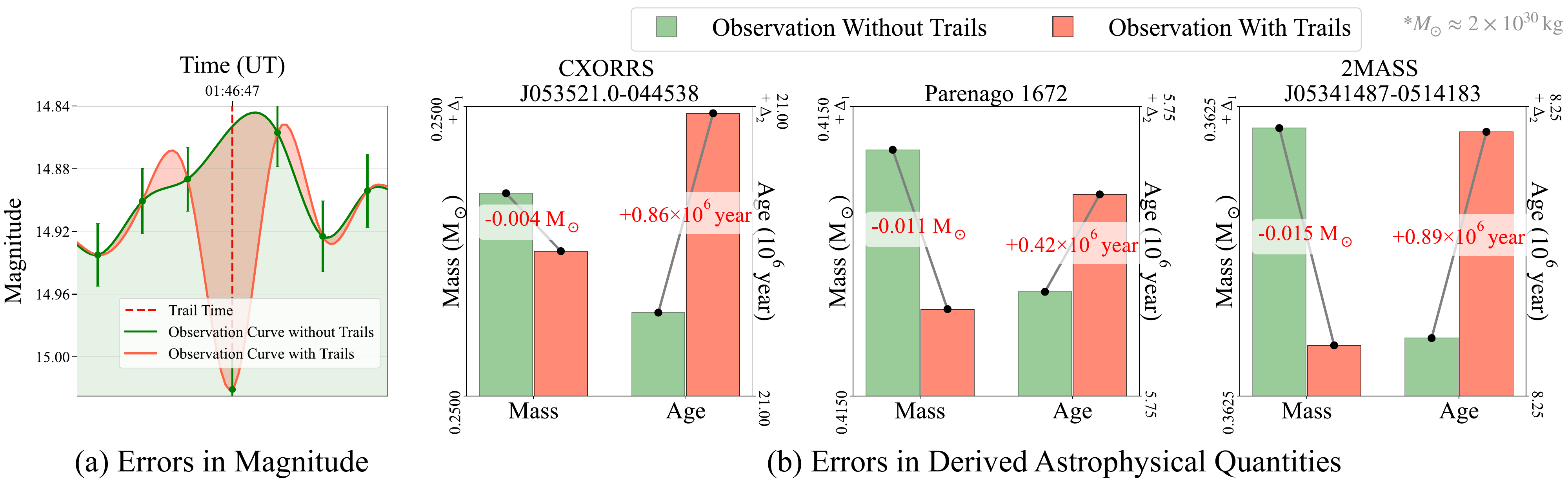}
\caption{Motivational study of the impact of satellite contamination. 
(a)~A satellite trail intersecting a star perturbs its measured brightness (magnitude). 
(b)~The resulting bias leads to errors in derived stellar properties (e.g., age and mass). 
}
\vspace{-1mm}
\label{fig:moti:whole}
\end{figure}

\section{Motivation}
\label{sec:moti}
\subsection{Scientific Impact of Satellite Trail Contamination}

In this section, we investigate the detrimental impact of satellite trail contamination on astronomical research. 
Satellite trails corrupt raw observation data, leading to systematic errors in scientific measurements. 
We illustrate the risks by presenting real-world cases of degradation induced by satellite trails.

The primary threat of satellite trails is the corruption of stellar photometry.
A single contaminated exposure can inflate variability metrics or distort statistics (See \cref{fig:moti:whole}(a)).
One datapoint affected by a satellite trail distorts the observed trend.
This highlights that a single streak can impair the reliability of time-domain analyses. 
The findings call for robust mitigation of such trails.

Such errors in magnitude and brightness can propagate into significant inaccuracies in derived physical quantities (e.g., stellar mass and age). %
An actual demonstration is shown in \cref{fig:moti:whole}(b), with three 
real-world stellar sources~\cite{wenger2000simbad}.
For this illustrative analysis, we assume that the contaminated sources are main-sequence stars and use the main-sequence lifetime as an age proxy.
We propagate the magnitude bias using \(F \propto 10^{-0.4\Delta m}\), \(L \propto F\) for fixed distance, \(L \propto M^{3.5}\), and \(t_{\mathrm{MS}} \propto M/L\), with other properties fixed.
The results introduce substantial bias, up to 3.9\% deviation in mass, and 10\% deviation in age. 
Thus, satellite trails remain a major unresolved source of systematic error in stellar photometry.

\subsection{Limitations of Existing Line Detection Techniques}
\begin{figure}[t]
\centering
\includegraphics[width=\textwidth]{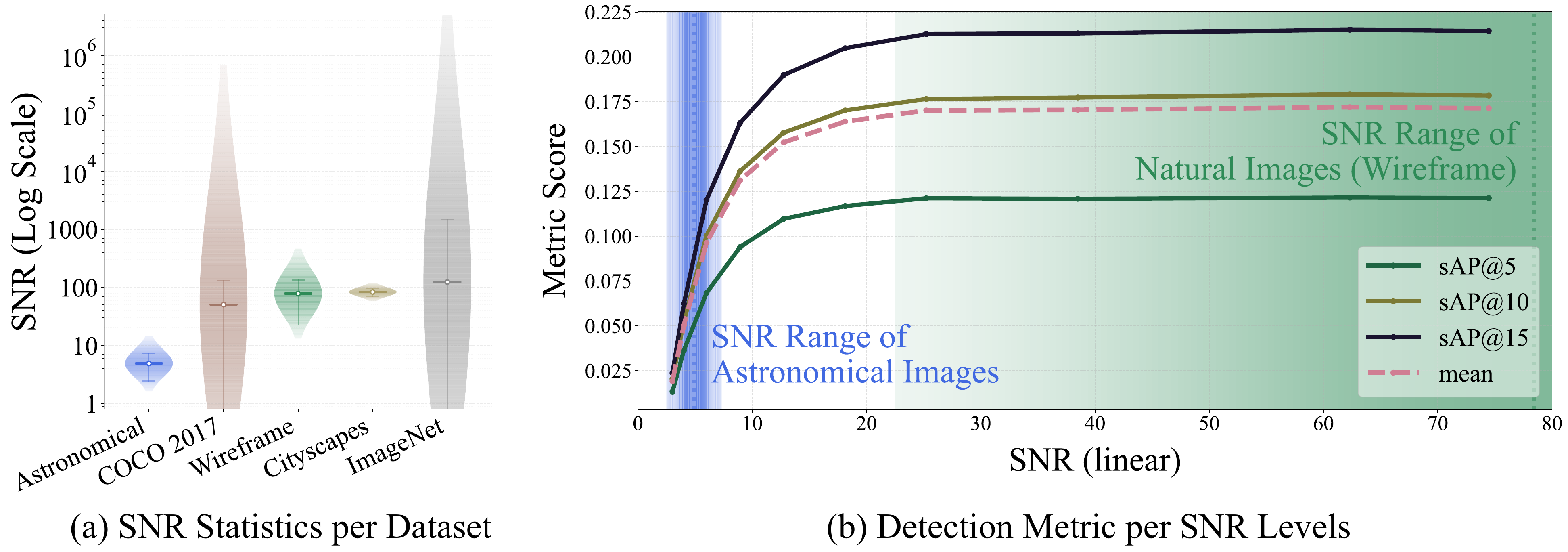}
\caption{(a) Visualization of SNR distribution per dataset. The astronomical images consists of very low SNR, compared to general domain datasets. (b) Line detection performance under varying signal-to-noise ratios (SNR) of images. 
Deep learning detectors 
degrade rapidly in the low-SNR region. The metric is structural Average Precision (sAP), which measures the accuracy of predicted lines against the ground truth. @5, @10, and @15 denote the pixel tolerance for matching prediction and ground-truth.}
\label{fig:moti:snr}
\end{figure}

To understand the unique challenge of satellite trail detection, we identify the fundamental gap between astronomical and natural imagery.
We characterize this gap using the signal-to-noise ratio (SNR), which quantifies the strength of meaningful signals relative to background noise. 
We further demonstrate that this discrepancy limits existing line detection methods for astronomical images.

From \cref{fig:moti:snr}(a), we observe a significant disparity in SNR between two image domains: astronomical images exhibit much lower SNR, compared to images from general vision datasets. 
While astronomical images exhibit SNR values of at most 16.84, general image datasets show median SNR values ranging from 35 to 116.
Unlike ordinary photography, which receives abundant photon flux, astronomical observations operate in a photon-starved regime where faint signals are often dominated by instrumental and atmospheric noise.
As a result, astronomical survey images operate in an inherently low-SNR regime. %

This observation sheds light on how linear structures exist in each domain.
In natural images, lines mostly exist as high-contrast boundaries between adjacent objects.
Satellite trails, however, are isolated features independent of stellar objects or noisy backgrounds. 
Without surrounding contextual cues or strong contrast, these trails are faint and extremely thin signals embedded within noise.

Such features counter the assumptions underlying existing line detection methods, which are devised for high-contrast edges and rich contextual information.
To demonstrate this limitation, we evaluate DeepLSD~\cite{deeplsd}, a representative learning-based line detector for natural images, on Wireframe images~\cite{huang2018learning} while progressively adding synthetic noise to reduce their SNR to astronomical levels. 
As shown in \cref{fig:moti:snr}(b), detection performance remains stable in the high-SNR regime typical of natural images (SNR > 70), but degrades rapidly as the SNR approaches that of astronomical imagery. 
In the single-digit SNR regime, detection metrics drop sharply, confirming that conventional vision-based methods are ill-suited for detecting faint linear structures under astronomical noise and motivating the need for a tailored approach. 
A visualized example of this noise-degradation process is presented in \cref{fig:snr_example} at \cref{app:snr}.

\section{Related Work}

\subsection{Classical Methods for Line Detection}
There were several early attempts %
to detect line structures in images through mathematical heuristics. 
The Hough~\cite{hough} and Radon~\cite{radon} transforms reformulate the task as identifying local peaks by mapping colinear pixels to a point.
The LSD~\cite{lsd}, ED~\cite{ed}, EDLines~\cite{edlines}, AG3line~\cite{ag3line}, and ELSED~\cite{elsed} construct line segments faster by aggregating pixels with consistent gradients from local anchors.
These detectors are often paired with the Canny edge detector~\cite{canny}. %
They often lack flexibility when pixel-intensity variations are too subtle to elicit stable responses, thus requiring extensive parameter tuning. 

\subsection{Deep Learning Approaches for Line Detection}
To overcome the limitations of prior handcrafted heuristics, deep learning techniques have been integrated into line detection.
Such methods often reframe the line segment detection problem into alternative learning tasks, such as region coloring~\cite{xue2019learning} or graph inference~\cite{ppgnet} by constructing graphs for detected junction points in an image. 
Another strategy~\cite {lightCNN_hough, deephough} is to apply the Hough transform to CNN features, training the model in the parameter space. %
DeepLSD~\cite{deeplsd} refines line segment predictions by leveraging deep image gradients to produce accurate endpoints and orientations.
DT-LSD~\cite{dt-lsd} introduces deformable transformers that adapt receptive fields to capture geometric variations in line structure.
Most learned line detectors are designed for general-domain images, where lines typically correspond to object boundaries with rich contextual information.

\subsection{Automated Satellite Trail Detection in Astronomical Images}
To address the challenges posed by the unique properties of astronomical trails, several works~\cite{asta, pyradon, astride, deepstreaks, ali2006satellite, streakML, unetLSD, maximask, yolosatrail} have focused on detecting satellite trails in astronomical survey. 
Early techniques %
leveraged geometric algorithms such as boundary-tracing~\cite{astride}, and morphological filtering~\cite{ali2006satellite} to capture and remove trails from telescope exposures.
Several methods~\cite{deepstreaks, streakML, yolosatrail} train CNN models or ML-based classifiers to identify satellite trails. %
More recent approaches, such as ASTA, MaxiMask, UNetLSD~\cite{maximask, asta, unetLSD}, combine segmentation networks with geometric detectors or multi-class contaminant labels to localize trails at the pixel level.

For model training, these methods require pixel-level trail annotations as supervision. 
However, obtaining annotations relies on expensive manual examination.
This motivates the necessity for approaches that can train trail detectors without such expensive pixel-level supervision.

\begin{figure}[t]
  \centering
  \includegraphics[width=\linewidth]{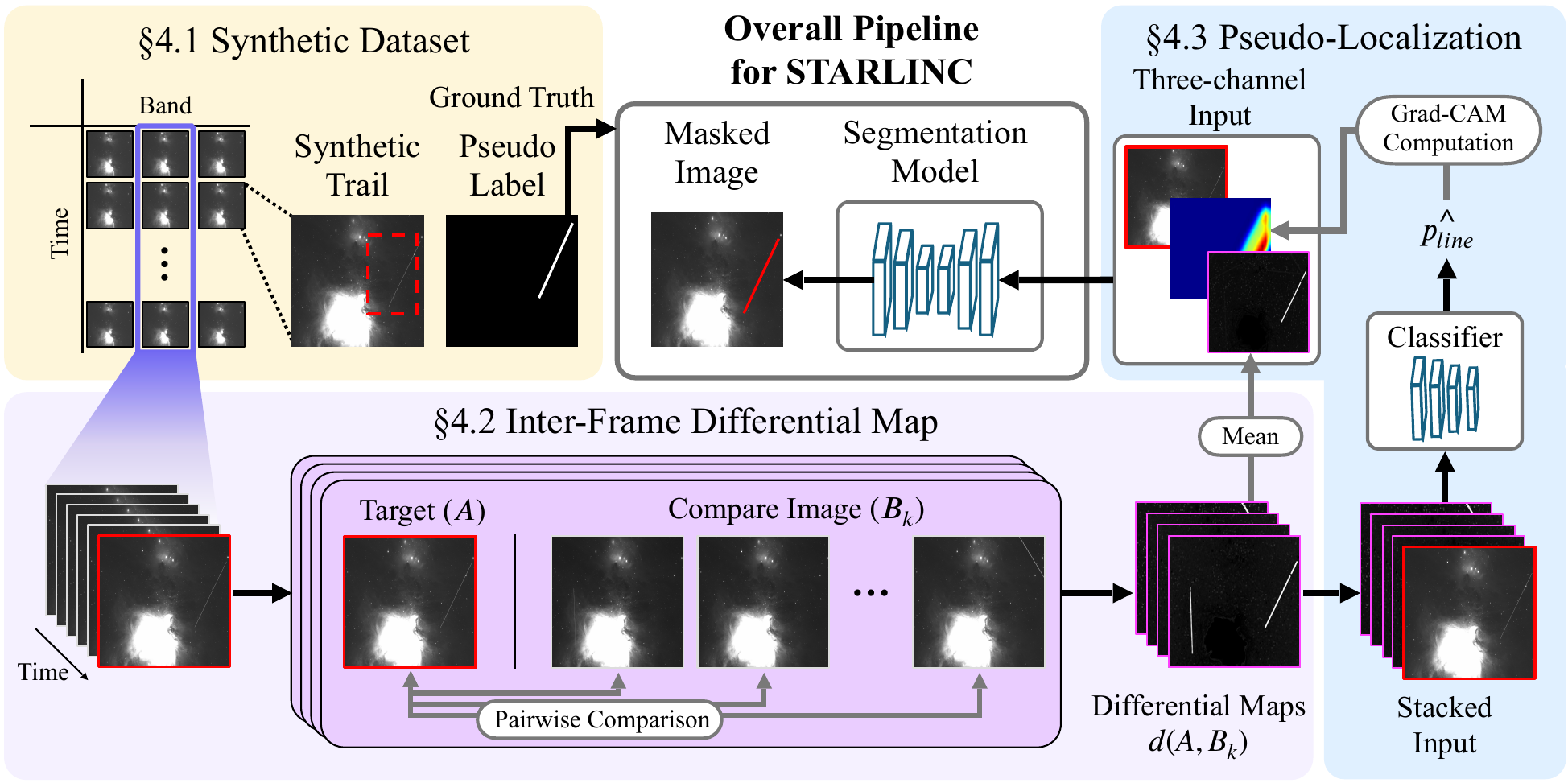}
  
  \caption{\textbf{Overall framework of \name.} %
  The pipeline consists of three stages.
  {Synthetic data generation (\S~\ref{sec:method:synth})}: Artificial satellite trails are injected into non-trail images to generate image-label pairs for segmentation model training. 
  Inter-frame differential map (\S~\ref{sec:method:classification}): Leveraging repeated exposures of the same sky region, \name suppresses static celestial structures and emphasizes transient satellite trails using locally aggregated temporal differences.
  Pseudo-localization via \gradcams (\S~\ref{sec:method:gradcam}): We convert an image-level trail classifier into a spatial cue by extracting Grad-CAM activation heatmaps, which provide coarse pseudo-localization of the trail and guide the downstream segmentation network without pixel-level annotations.
  }
  \label{fig:fig3}
\end{figure}

\section{Method}

Our framework (\cref{fig:fig3}) comprises four components that progressively generate pixel-level masks of satellite trails while preserving uncontaminated image regions.
To address the lack of pixel-level annotation and class imbalance problem, 
we first generate synthetic training samples by injecting simulated satellite trails into clean astronomical images (\S~\ref{sec:method:synth}). 
We then exploit temporal redundancy in repeated observations by computing inter-frame differential maps across adjacent exposures, which suppress persistent celestial backgrounds and emphasize transient satellite artifacts (\S~\ref{sec:method:classification}).
These inter-frame representations are used to obtain heatmaps that provide pseudo-localization cues for the later segmentation stage, which are extracted from a ResNet-based classifier trained to predict trail presence (\S~\ref{sec:method:gradcam}).
Finally, a U-Net-based~\cite{unet} segmentation model takes the original image, the corresponding inter-frame differential map, and the \gradcam to produce pixel-level masks without requiring manual annotations (\S~\ref{sec:method:masking}).

\subsection{Synthetic Trail Generation for Training} %
\label{sec:method:synth}

We introduce a synthetic trail generation strategy tailored to astronomical survey to address two limitations of real survey data. 
First, large-scale datasets typically lack pixel-level trail annotations, and off-the-shelf line detectors generalize poorly to astronomical survey. 
Second, the data are highly imbalanced, with only about 20\% of images containing satellite trails. 
To address both issues, we generate synthetic trails without manual masks for scalable training and for controlled sampling of trail occurrences.

We generate synthetic training samples by injecting simulated satellite trails into clean astronomical images. 
The resulting synthetic data provide paired images and pixel-level ground-truth masks, enabling direct training of the segmentation model without manual pixel-level annotations.

To maintain visual authenticity and physical fidelity, %
each synthetic trail is parameterized by four attributes: length $L$, orientation $\theta$, thickness $\tau$, and brightness $m$. 
The trail length, orientation, thickness, and brightness are sampled as
\begin{align}
L &\sim \mathrm{clip}\!\left(\mathcal{N}(D,\sigma_L^2);\alpha,\beta\right),\\
\theta &\sim \mathcal{U}(0,2\pi),\\
\tau &\sim \mathrm{B}(a,b),\\
m &\sim \mathcal{U}(m_{\min},m_{\max}).
\end{align}
Here, $\mathcal{N}(\mu,\sigma^2)$ denotes a normal distribution with mean $\mu$ and variance $\sigma^2$, $\mathcal{U}(u_{\min},u_{\max})$ denotes a uniform distribution over $[u_{\min},u_{\max}]$, and $\mathrm{B}(a,b)$ denotes a beta distribution with shape parameters $a$ and $b$. 

The trail position is sampled uniformly over the image plane. 
To better reflect real astronomical observations, the trail thickness $\tau$ is drawn from a Beta distribution that favors smaller values. 
This bias increases the frequency of thin trails in the synthetic data, which is important because satellite trails often appear as faint and narrow structures under low signal-to-noise conditions. 
By emphasizing such thin trails during training, the model learns to detect subtle linear artifacts that are difficult to distinguish from background noise.

\subsection{Inter-Frame Differential Map for Background Suppression}

\label{sec:method:classification}

\name leverages a key property of time-domain astronomical surveys: the same sky region is repeatedly imaged under nearly identical conditions. 
As a result, static celestial structures (e.g., stars and galaxies) remain highly consistent while transient artifacts such as satellite trails appear only in a subset of frames. 
We explicitly capture these temporal cues using inter-frame differential maps.

Given two normalized grayscale images $x_i, x_j \in \mathbb{R}^{H \times W} $, we construct a pixel-wise differential map $M_{i,j} \in \mathbb{R}^{H \times W}$. 
Rather than computing a purely point-wise distance, we aggregate local structural differences within a sliding window:
\begin{align}
M_{i,j}(m,n) = \sum_{(u,v)\in\mathcal{N}_k(m,n)}d(x_i(u,v),\, x_j(u,v)),
\label{eq:distance}
\end{align}
where $d(\cdot,\cdot)$ denotes a distance function and $\mathcal{N}_k(m,n)$ is a $k \times k$ neighborhood of pixel $(m,n)$.
This local aggregation enables the differential map to capture regional variations rather than pixel noise, thereby improving robustness.

An important consideration in selecting $d(\cdot,\cdot)$ is robustness to inherent observational noise of ground-based surveys. 
To this end, we adopt the Structural Similarity Index (SSIM) to define the distance as its complement:
\begin{align}
d(x_i, x_j) = 1 - \mathrm{SSIM}(x_i, x_j) = 1 - \frac{(2\mu_i \mu_j + C_1)(2\sigma_{ij} + C_2)}{(\mu_i^2 + \mu_j^2 + C_1)(\sigma_i^2 + \sigma_j^2 + C_2)},
\label{eq:diff}
\end{align}
where $\mu_i$ and $\mu_j$ denote local means, $\sigma_i^2$ and $\sigma_j^2$ denote local variances, and $\sigma_{ij}$ denotes the local covariance computed within the window. 
$C_1$ and $C_2$ are small constants that stabilize the computation when local statistics are close to zero.
A visualized example of this process is presented in \cref{fig:method:ssim} at \cref{app:sec:input_ex}.

For each target frame, the mean differential map is computed by aggregating pairwise SSIM distances against the remaining frames. %
Since static background structures are shared across frames, they are largely suppressed despite natural observational noise. 
In contrast, a satellite trail appearing in only one frame induces a localized peak in the distance measure. %
These differential maps serve as the primary input representation for the subsequent segmentation stages (\S~\ref{sec:method:gradcam}).

\subsection{Pseudo-localization via Activation Maps}
\label{sec:method:gradcam}

Even with temporal differential maps, the segmentation stage benefits from a spatial prior indicating where the trail is likely to appear.
To provide a lightweight localization signal, we use an auxiliary trail-presence classifier as a pseudo-localizer.
The resulting gradient-based activation maps~\cite{gradcam} highlight the regions most responsible for the trail prediction and guide the segmentation network.

We train a classifier that takes input of the original grayscale frames and their corresponding differential maps. 
The classifier outputs a probability of trail presence, $\hat{p}_{\mathrm{line}}$.
To obtain a spatial localization cue from the trained classifier, we extract activation maps using Grad-CAM~\cite{gradcam}. 
Let $\{f^k\}_{k=1}^{K}$ denote the $K$ feature maps of the final convolutional layer, and let $y_c$ denote the logit for the trail-present class. 
We compute channel-wise importance weights by globally averaging the gradient of $y_c$ with respect to each feature map.
\begin{equation}\label{eq:gradcam_weights}
  \alpha_k
  = \frac{1}{HW}\sum_{i=1}^{H}\sum_{j=1}^{W}
    \frac{\partial y_c}{\partial f^k(i,j)}.
\end{equation}

These weights are then used to produce a coarse localization heatmap via a weighted combination followed by a ReLU activation,
\begin{equation}\label{eq:gradcam_heatmap}
  H_{\mathrm{GC}}
  = \mathrm{ReLU}\!\left(\sum_{k=1}^{K}\alpha_k\,f^k\right).
\end{equation}
The resulting heatmap $H_{\mathrm{GC}}\in\mathbb{R}^{H'\times W'}$ highlights spatial regions %
that contribute most to trail prediction and are directly fed into the segmentation network (\S~\ref{sec:method:masking}), without requiring pixel-level supervision.

\subsection{Overall Pipeline for Trail Masking}
\label{sec:method:masking}

The final stage of \name produces pixel-level binary trail masks using a segmentation network trained on synthetic supervision from \S~\ref{sec:method:synth}. 
For each target image, the segmentation network receives a three-channel input consisting of the original grayscale image $I$,
the mean inter-frame differential map $\bar{M}$, and the Grad-CAM heatmap
$H_{\mathrm{GC}}$:
\begin{align}
    X = [I, \bar{M}, H_{\mathrm{GC}}] \in \mathbb{R}^{H \times W \times 3}.
\end{align}
Given this composite input, the U-Net segmentation network $f_{\theta}$ predicts a single-channel trail probability map,
\begin{align}
    \hat{Y} = f_{\theta}(X) \in [0,1]^{H \times W},
\end{align}
which is thresholded at inference time to obtain the final binary trail mask.

\noindent\textbf{Training.}
During training, \name constructs the same three-channel representation that is used at inference. 
We first select a subset of target images within each temporal group and inject synthetic trails into those images. Each temporal group contains 30 repeated exposures of the same sky field. 
For a synthetic target image $I^{\mathrm{syn}}$, the mean differential map $\bar{M}^{\mathrm{syn}}$ is computed by comparing the target image against the remaining exposures in the group using the procedure in \S~\ref{sec:method:classification}. 
The trail-presence classifier then produces a Grad-CAM heatmap $H_{\mathrm{GC}}^{\mathrm{syn}}$ as described in \S~\ref{sec:method:gradcam}.
The resulting training input is
\begin{align}
    X^{\mathrm{syn}}
    =
    [I^{\mathrm{syn}}, \bar{M}^{\mathrm{syn}},
    H_{\mathrm{GC}}^{\mathrm{syn}}]
    \in \mathbb{R}^{H \times W \times 3}.
\end{align}
The segmentation network predicts $\hat{Y}^{\mathrm{syn}} = f_{\theta}(X^{\mathrm{syn}})$ and is supervised by the known synthetic binary trail mask $Y^{\mathrm{syn}}$. 
Thus, pixel-level supervision is obtained without manually annotated real trail masks.

\noindent\textbf{Inference.}
Given a target image $I$ from a temporal group of repeated exposures, \name computes the two auxiliary inputs: the mean differential map $\bar{M}$ and the classifier heatmap $H_{\mathrm{GC}}$. 
The trained U-Net predicts
\begin{align}
    \hat{Y} = f_{\theta}([I, \bar{M}, H_{\mathrm{GC}}]),
\end{align}
and converts the probability map into a binary satellite-trail mask. 
The predicted mask can be applied to the original astronomical image to exclude contaminated pixels while retaining uncontaminated regions.

\section{Experiments}

\subsection{Experimental Settings}
\label{sec:exp_setting}

To evaluate the performance of \name on real astronomical observations, we construct the primary dataset from 7DT~\cite{7dt} observations of the Orion Molecular Clouds region. 
The 7DT project is a wide-field ground-based survey that repeatedly observes the same sky field, providing multi-epoch exposures.
Temporally adjacent frames are used to construct inter-frame differential maps with window size $k=5$, and we evaluate our method on classification, segmentation, and trail removal tasks. 
The dataset contains 1,140 samples, which we divided into 720 training, 180 validation, and 240 test samples.
For synthetic trail generation, the parameters in Eq.~(1)–(4) are set as 
$D = 0.5\min(H,W)$, $\sigma_L = 0.25D$, $\alpha = 0.1D$, $\beta = 0.8D$, 
$(a,b) = (2,6)$, and $(m_{\min}, m_{\max}) = (0.35\cdot255,\,0.65\cdot255)$.
More detailed experiment settings are included in appendix \cref{app:sec:exp_setting}.

\subsection{Evaluation on Satellite Trail Segmentation Task}

\begin{table}[t]
\centering
\caption{Segmentation performance comparison.}
\label{tab:seg_perf}
\centering
\resizebox{.97\textwidth}{!}{%
\setlength{\tabcolsep}{5pt}
\begin{tabular}{llcccccc}
\toprule
\multirow{3}{*}{Category} & \multirow{3}{*}{Method} & \multicolumn{6}{c}{Metrics} \\
\cmidrule(lr){3-8}
& & mIoU & Dice & Prec. & Recall & \makecell{ROC\\AUC} & \makecell{PR\\AUC}\\
\midrule
\multirow{3}{*}{Classical}

	&	Hough~\cite{hough}	&	0.142	&	0.202	&	0.785	&	0.293	& 0.692 & 0.741 \\
	&	Radon~\cite{radon}	&	0.000	&	0.001	&	0.653	&	0.001	& 0.670 & 0.696 \\
	&	LSD~\cite{lsd}	&	0.074	&	0.126	&	0.077	&	0.517	& 0.776 & 0.714\\
	\midrule										
	\multirow{2}{*}{Learning-based}										
	&	DeepLSD~\cite{deeplsd}	&	0.116	&	0.192	&	0.120	&	0.642	& 0.867 & 0.827 \\
	&	DT-LSD~\cite{dt-lsd}	&	0.240	&	0.332	&	0.663	&	0.507	& 0.769 & 0.838\\
	\midrule										
	\multirow{3}{*}{Trail detection}										
	&	ASTA~\cite{asta} 	&	0.148	&	0.217	&	0.399	&	0.236	& 0.730 & 0.511\\
	&	MaxiMask~\cite{maximask}	&	0.155	&	0.268	&	0.838	&	0.159 & 0.795 & 0.830	\\
    &	UnetLSD~\cite{unetLSD}	&	0.197	&	0.329	&	0.303& 0.359	&	0.622 & 0.399	\\
	\midrule 
    \rowcolor{blue!8}
    Proposed	&	\textbf{STARLINC}	&	0.485	&	0.653	&	0.815	& 0.545	& 0.918	& 0.865 \\
														
\bottomrule
\end{tabular}
}
\end{table}

In this section, we evaluate the satellite trail segmentation ability of \name by measuring pixel-level localization. \Cref{tab:seg_perf} reports mIoU, Dice, Precision, Recall, ROC-AUC, and PR-AUC across classical line detectors, learning-based line detectors, and trail detection methods, together with our method.
All metrics are computed against ground-truth masks manually annotated using LabelMe~\cite{russell2008labelme}.

\name outperforms all baselines on overlap-based metrics (mIoU/Dice), suggesting substantially more spatially coherent masks. 
Notably, several baselines show non-trivial ROC-AUC/PR-AUC yet still fail to yield meaningful pixel-level overlap, indicating that discrimination metrics alone do not imply accurate localization. 
For example, Hough attains high precision (0.785) but misses much of the trail extent (recall 0.293), resulting in limited overlap (mIoU 0.142), while Radon and LSD provide little to no usable masks (mIoU 0.000/0.074).

\begin{figure}[t]
\centering
\includegraphics[width=\linewidth]{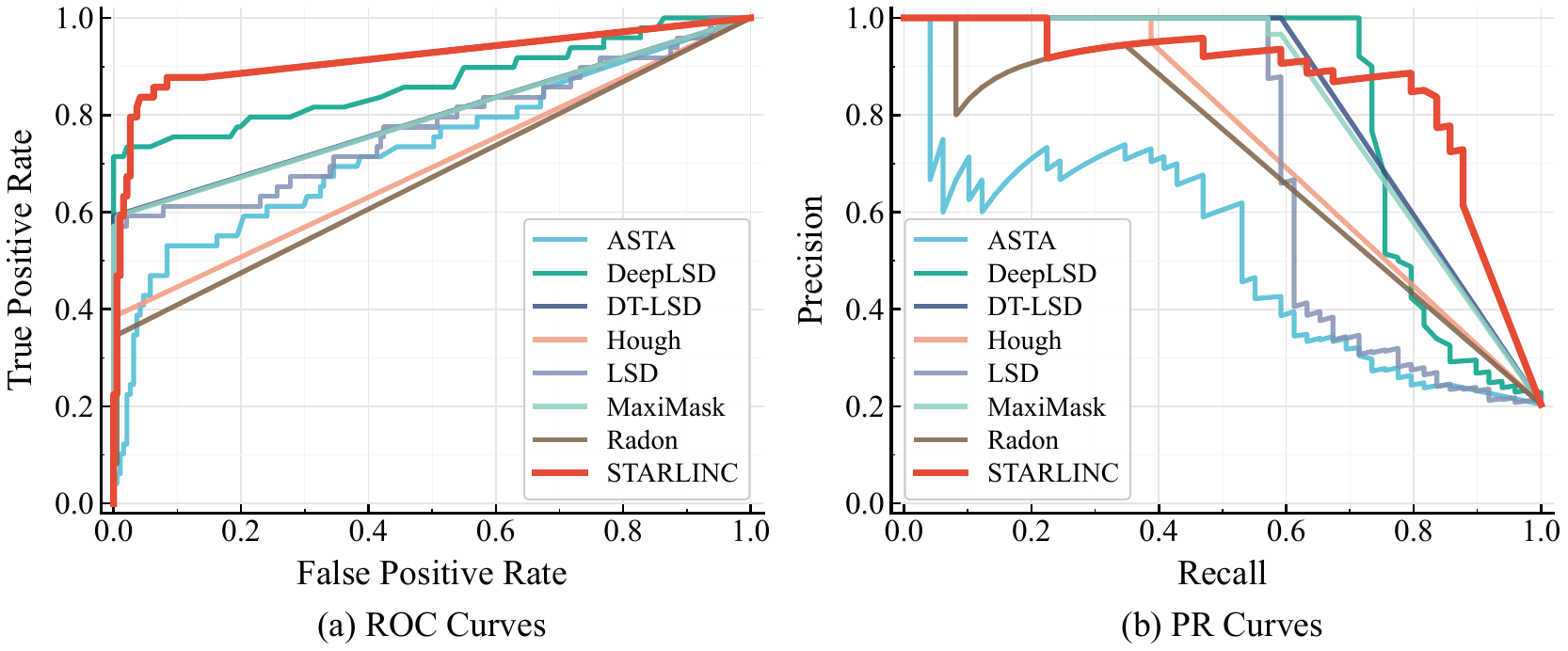}

\caption{Performance evaluation for segmentation model using (a) ROC curve and (b) PR curve showing performance across different decision thresholds.}

\label{fig:exp:roc_pr_seg}

\end{figure}
Among learning-based baselines, DeepLSD achieves the highest recall (0.642) but extremely low precision (0.120), consistent with overly dense predictions.
While DT-LSD provides a better balance (mIoU 0.240, Dice 0.332) than DeepLSD, it remains far behind \name. 
The other methods underperform: ASTA yields low scores across metrics, and MaxiMask shows a mismatch between image-level recognition and localization, with high precision (0.838) but low recall (0.159), leading to limited overlap (mIoU 0.155, Dice 0.268).
In contrast, \name achieves the best localization (mIoU 0.485, Dice 0.653) with high precision (0.815) and strong recall (0.545), doubling the best baseline mIoU.

In addition, \cref{fig:exp:roc_pr_seg} visualizes the ROC and PR curves whose trends match the AUC metrics in \cref{tab:seg_perf}. 
In particular, \name maintains a more favorable trade-off, with higher ROC-AUC (0.918) and PR-AUC (0.865) than all baselines. 
This agreement further supports that the gains are not confined to a single operating point, but persist across a broad range of decision thresholds.

\begin{figure}[t]
  \centering
  \includegraphics[width=\textwidth]{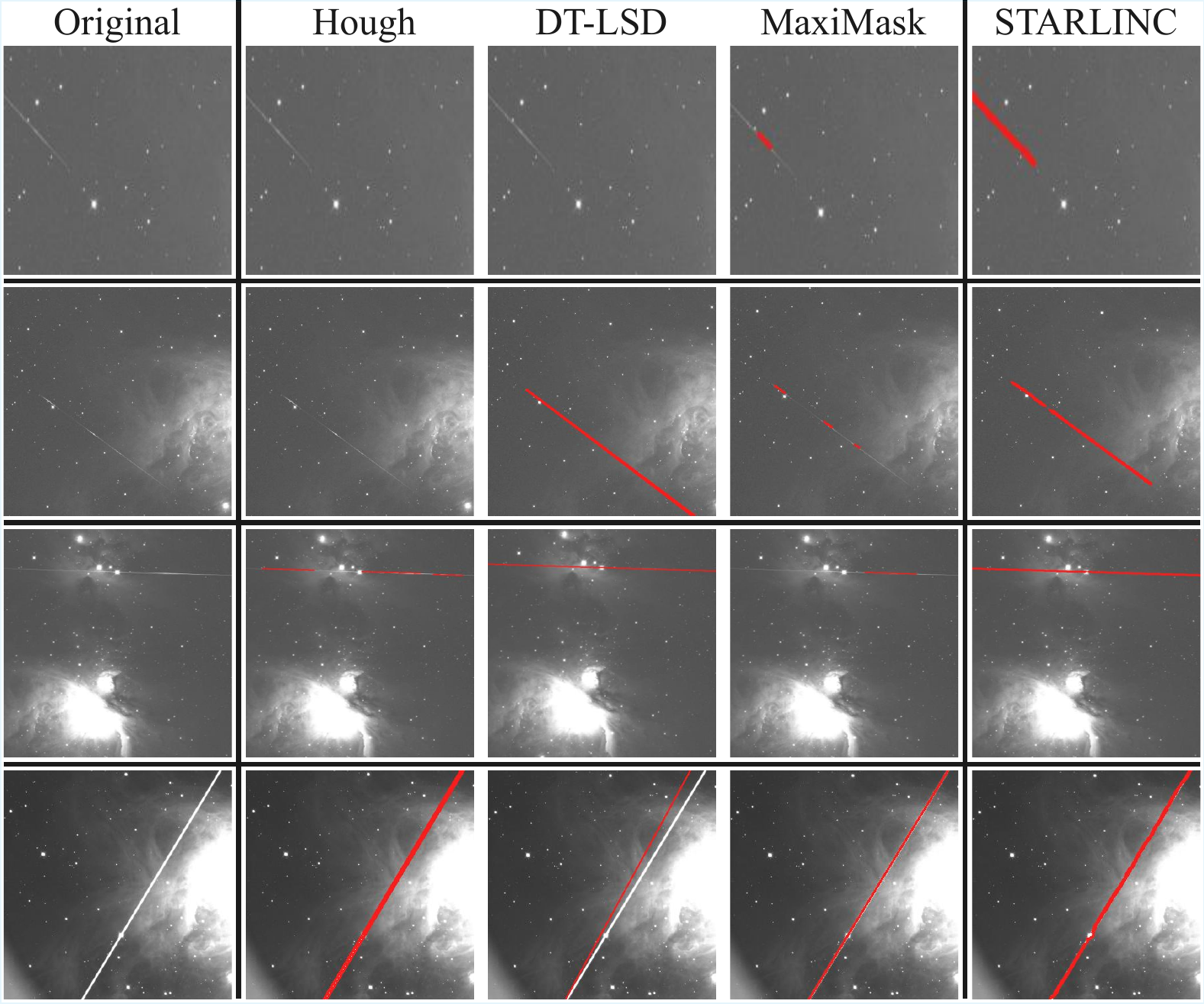}
  \caption{\textbf{Qualitative examples of satellite trail masking.}
  Each row shows a different astronomical image containing a satellite trail.
  From left to right, the original observation, and trail masks predicted by Hough~\cite{hough}, DT-LSD~\cite{dt-lsd}, MaxiMask~\cite{maximask}, and \name, where detected trail locations are overlaid in red. If the model fails to detect the line, the output remains identical to the original image.}
  \label{fig:fig6}
\end{figure}

\subsection{Qualitative Analysis of Satellite Trail Segmentation}

Qualitative results are shown in \cref{fig:fig6}.
The Hough transform produces short, fragmented mask segments, while DT-LSD outputs thicker, more dispersed masks than the trail.
MaxiMask produces no visible mask in most cases, consistent with its near-zero quantitative scores.
In contrast, \name produces continuous masks that closely follow the true trail geometry. %
This performance is enabled by the three-channel input design, where the mean differential map and \gradcam encode temporal structure, allowing a model trained on synthetic data to generalize effectively to real observations.
These results demonstrate that reliable pixel-level segmentation can be achieved by directly extending image-level classification with synthetic supervision, without requiring costly manual annotation of real astronomical survey. 
A comparison among all models is provided in \cref{app:sec:qual_results}.

\subsection{Cross-Dataset Validation}
\label{sec:cross_dataset_validation}

In this section, we evaluate the cross-dataset generalizability of STARLINC. 
For this evaluation, we use two additional datasets: an NGC-region dataset captured with the 7DT telescope~\cite{7dt} and a publicly available ZTF dataset~\cite{graham2019zwicky}.
The NGC-region dataset contains 4,812 images, and the ZTF dataset contains 4,875 images. 
To construct evaluation ground truth, human annotators manually masked the satellite trails in these datasets.
The \name model is trained only on the original Orion Molecular Clouds region 7DT training set and directly applied to both target datasets without target-dataset fine-tuning.
For learning-based baselines, we report their fine-tuned settings when
applicable, denoted by $^\dagger$.

\begin{table}[t]
\centering
\caption{
Cross-dataset evaluation on ZTF and NGC-region observations.
}
\label{tab:cross_dataset}
\resizebox{\linewidth}{!}{
\begin{tabular}{llcccccccc>{\columncolor{blue!8}}c}
\toprule
\multirow{2}{*}{Dataset} & \multirow{2}{*}{Metric} & \multicolumn{9}{c}{Method} \\
\cmidrule(lr){3-11}
& & Hough & Radon & LSD & DeepLSD$^\dagger$ & DT-LSD$^\dagger$ & ASTA$^\dagger$ & MaxiMask$^\dagger$ & UnetLSD$^\dagger$ & \textbf{\name} \\
\midrule
\multirow{2}{*}{ZTF}
& mIoU & 0.166 & 0.029 & 0.288 & 0.321 & 0.019 & 0.067 & 0.365 & 0.163 & \textbf{0.469} \\
& Dice & 0.285 & 0.057 & 0.447 & 0.486 & 0.038 & 0.126 & 0.535 & 0.280 & \textbf{0.639} \\
\midrule
\multirow{2}{*}{NGC}
& mIoU & 0.115 & 0.001 & 0.050 & 0.199 & 0.011 & 0.020 & 0.008 & 0.120 & \textbf{0.471} \\
& Dice & 0.206 & 0.002 & 0.095 & 0.331 & 0.021 & 0.040 & 0.015 & 0.215 & \textbf{0.640} \\
\bottomrule
\end{tabular}
}
\end{table}

As shown in Table~\ref{tab:cross_dataset}, \name achieves the best mIoU and Dice on both datasets,
demonstrating that \name generalizes well beyond the original Orion Molecular Clouds region. 
Notably, \name leverages inter-frame differential maps that emphasize frame-specific satellite trails while suppressing persistent celestial structures, 
allowing the learned representation to transfer effectively across different observational fields and surveys.

\section{Analysis}
\subsection{Ablation Study}

\begin{minipage}[]{0.54\textwidth}
We perform an ablation study to analyze the contribution of the key components in the segmentation pipeline.
\cref{tab:abl_seg_full} reports the effect of the inter-frame differential map (\S~\ref{sec:method:classification}) and the pseudo-localization via \gradcam(\S~\ref{sec:method:gradcam}).
When both components are removed, the model fails to produce meaningful masks, resulting in zero
\end{minipage}
\hfill
\begin{minipage}{0.43\linewidth}
    \vspace{-4mm}
    \centering
    \captionof{table}{Ablation results of segmentation task.}
    
\centering

\label{tab:abl_seg_full}
\setlength{\tabcolsep}{4.6pt}
\resizebox{\linewidth}{!}{%
\begin{tabular}{cccccc}
\toprule
\multicolumn{2}{c}{Method} & \multicolumn{4}{c}{Metrics} \\
\cmidrule(lr){1-2}\cmidrule(lr){3-6}
 \S~\ref{sec:method:classification} & \S~\ref{sec:method:gradcam} & mIoU & Dice & Prec. & Recall \\
\midrule

\xmark & \xmark &	0.000	&	0.000	&	1.000	&	0.000	\\
\cmark & \xmark &	0.410	&	0.582	&	0.585	&	0.578	\\
\xmark & \cmark &	0.198	&	0.331	&	0.276	&	0.411	\\
\cmark & \cmark &	0.485	&	0.653	&	0.815	&	0.545	\\

\bottomrule
\end{tabular}}

\end{minipage}
    \vspace{1.1mm}

 \noindent segmentation performance. \
Using only the inter-frame differential map already yields strong segmentation performance, and adding the heatmap further improves the results.
Their combination performs best because the differential map suppresses persistent celestial structures while the \gradcam guides trail localization.
Additional sensitivity analysis is provided in \cref{app:sec:synthetic_fraction}.

\begin{table*}[t]
\centering

\begin{minipage}[t]{0.26\textwidth}
\centering
\caption{Differential map comparison.}
\label{tab:abl_diff}
\vspace{-3.7mm}
\renewcommand{\arraystretch}{1.28}
\resizebox{\linewidth}{!}{%
\setlength{\tabcolsep}{3.6pt}
\begin{tabular}{lc}
\toprule
Method & mIoU \\
\midrule
RMSE & $0.469 \pm 0.014$ \\
MAE  & $0.472 \pm 0.002$ \\
ZOGY & $0.392 \pm 0.016$ \\
OIS  & $0.431 \pm 0.033$ \\
DIA  & $0.436 \pm 0.008$ \\
\rowcolor{blue!8}
SSIM & $0.485 \pm 0.012$ \\
\bottomrule
\end{tabular}}
\end{minipage}
\hfill
\begin{minipage}[t]{0.73\textwidth}
\centering
\caption{Analysis of Alternative Designs.}
\label{tab:alternative_designs}
\renewcommand{\arraystretch}{0.55}
\resizebox{\linewidth}{!}{%
\setlength{\tabcolsep}{3.4pt}
\begin{tabular}{llcccc}
\toprule
Category & Method & mIoU & Dice & Prec. & Recall \\
\midrule
Image-only segmentation
& SAM$^\dagger$~\cite{kirillov2023segment} & 0.358 & 0.528 & 0.449 & 0.640 \\
\midrule
Synthetic supervision
& PSF & 0.539 & 0.700 & 0.748 & 0.658 \\
\midrule
\multirow{3}{*}{\makecell[l]{Inter-frame\\representation}}
& Otsu & 0.045 & 0.004 & 0.002 & 0.761 \\
& Morph. & 0.158 & 0.270 & 0.172 & 0.630 \\
& Hough & 0.259 & 0.178 & 0.105 & 0.741 \\
\midrule
\multirow{2}{*}{Localization cue}
& IRNet$^\dagger$~\cite{ahn2019weakly} & 0.131 & 0.232 & 0.373 & 0.169 \\
& SEAM$^\dagger$~\cite{wang2020self} & 0.039 & 0.074 & 0.106 & 0.057 \\
\midrule
\rowcolor{blue!8}
Proposed
& \textbf{\name} & 0.485 & 0.653 & 0.815 & 0.545 \\
\bottomrule
\end{tabular}}
\end{minipage}

\end{table*}

\subsection{Comparison of Inter-Frame Differential Maps}
\label{sec:diff_map_comparison}

The inter-frame differential map is a central cue in \name, suppressing persistent celestial content while highlighting frame-specific satellite trails.
We compare the SSIM-based structural dissimilarity map with RMSE, MAE, and astronomy-oriented subtraction methods, including ZOGY~\cite{zackay2016proper}, OIS~\cite{alard1998method}, and DIA~\cite{bramich2008new}.
These methods are treated as differential-map alternatives rather than stand-alone segmentation baselines, since they produce difference images rather than binary masks.
All variants use the same segmentation architecture and training setup, with only the differential-map construction changed.

As shown in \cref{tab:abl_diff}, SSIM performs best among the differential maps.
RMSE and MAE are sensitive to brightness fluctuations and residual stellar cores, while ZOGY, OIS, and DIA are mainly optimized for point-source variability and normalized subtraction.
Since satellite trails are line-like outliers, SSIM better captures frame-specific structural changes than pixel-wise flux differences.
Thus, the SSIM-based map provides a stronger masking cue, while physical subtraction methods may still require tuning for extended artifacts.
Qualitative comparisons are provided in \cref{fig:diff}.

\subsection{Analysis of Alternative Designs}
\label{sec:additional_baselines}

The three-channel design of \name naturally suggests several alternatives.
Instead of combining the original image \(I\), inter-frame differential map \(\bar{M}\), and \gradcam \(H_{\mathrm{GC}}\), we consider whether each cue can be used alone or replaced by an existing technique. 
We organize these alternatives by their target cue in \cref{tab:alternative_designs}.

\paragraph{\textbf{Image-only segmentation.}}
A natural baseline is to apply a general-purpose segmentation model directly to the astronomical image.
We evaluate SAM~\cite{kirillov2023segment} as an image-only segmentation alternative, without using inter-frame differential maps or classifier heatmaps.
As shown in \cref{tab:alternative_designs}, SAM obtains 0.358 mIoU and 0.528 Dice, which is below \name.
This suggests that generic image segmentation alone is insufficient for low-SNR satellite trail masking, where trails are thin and lack the object-like appearance priors of natural images.

\paragraph{\textbf{Synthetic supervision.}}

We further examine whether improved synthetic-trail realism can enhance segmentation performance.
For this purpose, we replace the geometric trail renderer with a PSF-aware renderer while keeping the remaining \name pipeline unchanged.
As shown in \cref{tab:alternative_designs}, PSF-aware rendering improves mIoU and Dice from 0.485/0.653 to 0.539/0.700, suggesting that more realistic synthetic trails can improve mask overlap. 
However, its lower precision relative to the main \name configuration (0.748 compared with 0.815) suggests that these gains in overlap metrics are accompanied by reduced precision. 
We therefore include it as an additional analysis.%

\paragraph{\textbf{Inter-frame representation.}}

We assess whether the inter-frame differential map can serve as a direct basis for mask generation.
Specifically, we apply Otsu thresholding, morphological filtering, and Hough-based line extraction to the differential map. Although these methods achieve relatively high recall (\cref{tab:alternative_designs}), their precision remains very low because residual stellar cores, detector artifacts, and background fluctuations also produce strong responses.
This shows that the differential map is effective as a localization cue but requires a learned segmentation model to distinguish satellite trails from other transient structures.

\paragraph{\textbf{Localization cue.}}

Since the heatmap provides coarse localization of trail-like regions, we examine whether it can be used for weakly supervised segmentation.
We evaluate IRNet~\cite{ahn2019weakly} and SEAM~\cite{wang2020self}, which derive masks from classifier localization cues.
Both perform substantially worse than \name.
Their activation maps typically capture only the most discriminative trail fragments, not the full thin trail extent.
Instead, \name uses the heatmap only as an auxiliary cue while obtaining pixel-level supervision from synthetic trail masks.

\section{Conclusion}

Satellite trail contamination is becoming a persistent problem for modern astronomical surveys as low Earth orbit constellations expand. 
We presented \name, which leverages temporal redundancy across repeated observations to isolate transient satellite artifacts. 
Using inter-frame differential maps, synthetic trail generation, and activation-map-based pseudo-localization, \name enables pixel-level segmentation without manually annotated masks from real survey data. 
Experiments on the 7DT dataset show clear gains over classical and learning-based baselines, and cross-dataset evaluations further demonstrate that \name maintains strong performance across survey datasets, indicating strong generalizability. 
By selectively masking contaminated regions instead of discarding entire frames, \name offers a practical and scalable approach for handling satellite-induced artifacts in large-scale astronomical surveys.

\section*{Acknowledgements}
This work has been mainly supported by the Basic Research Laboratory Program through the National Research Foundation of Korea (NRF) funded by the Korea government (MSIT) (RS-2024-00416859). 
Jinho Lee is the corresponding author, and he is also funded by: 
RS-2024-00395134, %
RS-2024-00347394, %
RS-2026-25548502, %
RS-2026-25495605, %
RS-2025-00564840, %
RS-2023-00256081, %
RS-2026-25549926  %

\bibliographystyle{splncs04}
\bibliography{main}

\clearpage
\appendix

\begin{center}
{\Large\bfseries Appendix}\\[-0.2em]
\end{center}
\vspace{0.6em}

\section{Detailed Experiment Settings}
\label{app:sec:exp_setting}
We adopt a U-Net architecture with four encoder and decoder stages and three-channel inputs. 
The model is trained using the Adam optimizer with a batch size of 8 and an initial learning rate of $1\times10^{-4}$ for 200 epochs. 
Input images are padded to multiples of 32 to match the U-Net downsampling structure, and a threshold of 0.3 is applied during inference to obtain binary masks.
All experiments are implemented in Python 3.11 using PyTorch 2.1 with CUDA 12.8 and are conducted on NVIDIA A6000 and RTX 4090 GPUs.
During inference, \name runs at 0.543s/img with 2.25GiB peak GPU memory.

\section{Additional Cross-Dataset Generalization on Astronomical Observations}

\label{app:sec:other_data}

In this section, we evaluate the cross-dataset generalizability of \name. 
The model used in this experiment is trained on our dataset and directly applied to astronomical images from different observational fields without additional retraining.
\cref{fig:other_data} shows qualitative examples from two different astronomical fields, Serpens and Messier~81. 
The Serpens images are drawn from the 7DT dataset~\cite{7dt}, while the Messier~81 images are obtained from the IRSA archive~\cite{graham2019zwicky}.

Despite the differences in background structure and object distribution, \name successfully localizes satellite trails while preserving surrounding celestial sources. 
In the Serpens example, \name detects a very faint trail that is barely visible in the original image. 
In the Messier~81 example, the method reliably identifies multiple trails within a single frame, further demonstrating strong cross-dataset generalization to unseen fields and trail configurations.
These results suggest that \name is not restricted to a specific dataset or observational domain and can be effectively applied to diverse astronomical images.

\section{Qualitative Results of Inter-Frame Differential Maps and Heatmaps}
\label{app:sec:input_ex}

\begin{figure}[h]
  \centering
  \includegraphics[width=\textwidth]{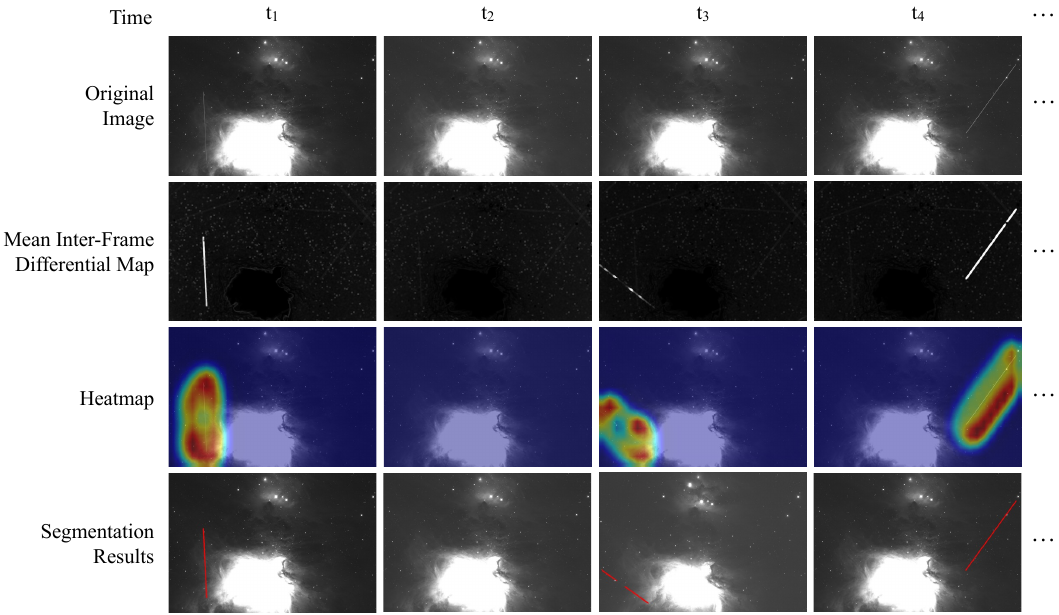}
  \caption{Visualization of the three input modalities used for segmentation over time: the original image, the mean inter-frame differential map~(\S~\ref{sec:method:classification}), and the heatmap~(\S~\ref{sec:method:gradcam}). Trail-containing frames (e.g., \(t_1\), \(t_3\), and \(t_4\)) show strong responses in both the differential map and the heatmap, while the trail-free frame (\(t_2\)) shows minimal activation.}
  \label{fig:method:ssim}
\end{figure}

In this section, we present qualitative examples of the three input modalities used by the segmentation model: the original image, the mean inter-frame differential map, and the heatmap.
\cref{fig:method:ssim} illustrates these three representations across consecutive frames.
The original image provides the full astronomical scene, including both static celestial objects and possible trail contamination. 
The mean inter-frame differential map (\S~\ref{sec:method:classification}) is obtained by averaging pairwise SSIM distances between the target frame and other frames in the same observation group.
This process emphasizes transient structures while suppressing static celestial objects that remain consistent across the group. 
The heatmap provides an additional localization cue derived from the classification model.

Frames containing satellite trails ($t_1$, $t_3$, $t_4$) exhibit a clear pattern.
In these cases, the trail produces a strong response in the differential map because it appears only in the target frame and is absent from the comparison frame, creating a localized structural inconsistency. 
The heatmap also highlights regions around the trail, indicating that the classifier attends to trail-contaminated areas. 
In contrast, a trail-free frame such as \(t_2\) exhibits only weak responses in both the differential map and the heatmap, indicating that both representations respond selectively to the presence of trails rather than to static background content.

These observations support our design choice for the segmentation input. 
The differential map captures temporal anomalies caused by moving satellite trails, while the heatmap provides complementary visual localization cues. 
Together, the differential map, heatmap, and original image form a complementary three-channel input for accurate trail segmentation.

\section{Further Qualitative Comparison}
\label{app:sec:qual_results}

In this section, we present additional qualitative comparisons as shown in \cref{fig:results_full2}.
While \cref{fig:fig6} provides representative cases of satellite trail removal, \cref{fig:results_full2} presents a larger set of examples, enabling evaluation across a wider variety of trail appearances and background conditions.

Classical line detection methods such as Hough, Radon, and LSD frequently fail to robustly capture satellite trails in astronomical images. 
These approaches often produce false detections on bright celestial structures and galaxy edges, while missing faint or partially visible trails. 
Learning-based line detectors, including DeepLSD and DT-LSD, demonstrate improved sensitivity in some cases but still produce fragmented detections or unstable responses, especially when trails are weak or blended with dense background structures.
Masking-based approaches such as ASTA and MaxiMask sometimes remove portions of the trail, but they often leave residual artifacts or remove excessive background regions, leading to excessive removal of valid astronomical data.

In contrast, \name consistently produces cleaner and more coherent trail localization and removal results. 
The proposed method effectively captures the full extent of satellite trails while preserving surrounding celestial structures, even in challenging scenarios involving faint trails, complex backgrounds, and 
trails of varying brightness.
These results highlight the robustness of \name for practical satellite trail removal in real astronomical survey.

\section{Classification-based Trail Detector Details}

To provide localization cues for segmentation, we train a classification model as described in \S~\ref{sec:method:gradcam} to generate heatmap-based representations. 
In this section, we report additional details and experimental results of the classifier.

\subsection{Architecture and Training Details}

For classification, we employ a ResNet-34 backbone trained using the Adam optimizer with an initial learning rate of $1\times10^{-4}$ and a batch size of 8 for 50 epochs. 
A ReduceLROnPlateau scheduler is applied with a reduction factor of 0.5 and patience of 5 epochs, with a minimum learning rate of $1\times10^{-7}$.

As described in \S~\ref{sec:method:synth}, synthetic satellite trails are injected into clean images to balance the training distribution. 
This augmentation mitigates the inherent class imbalance in satellite trail detection and facilitates stable training of the classification model.

\begin{figure}[h]
\centering
\includegraphics[width=\linewidth]{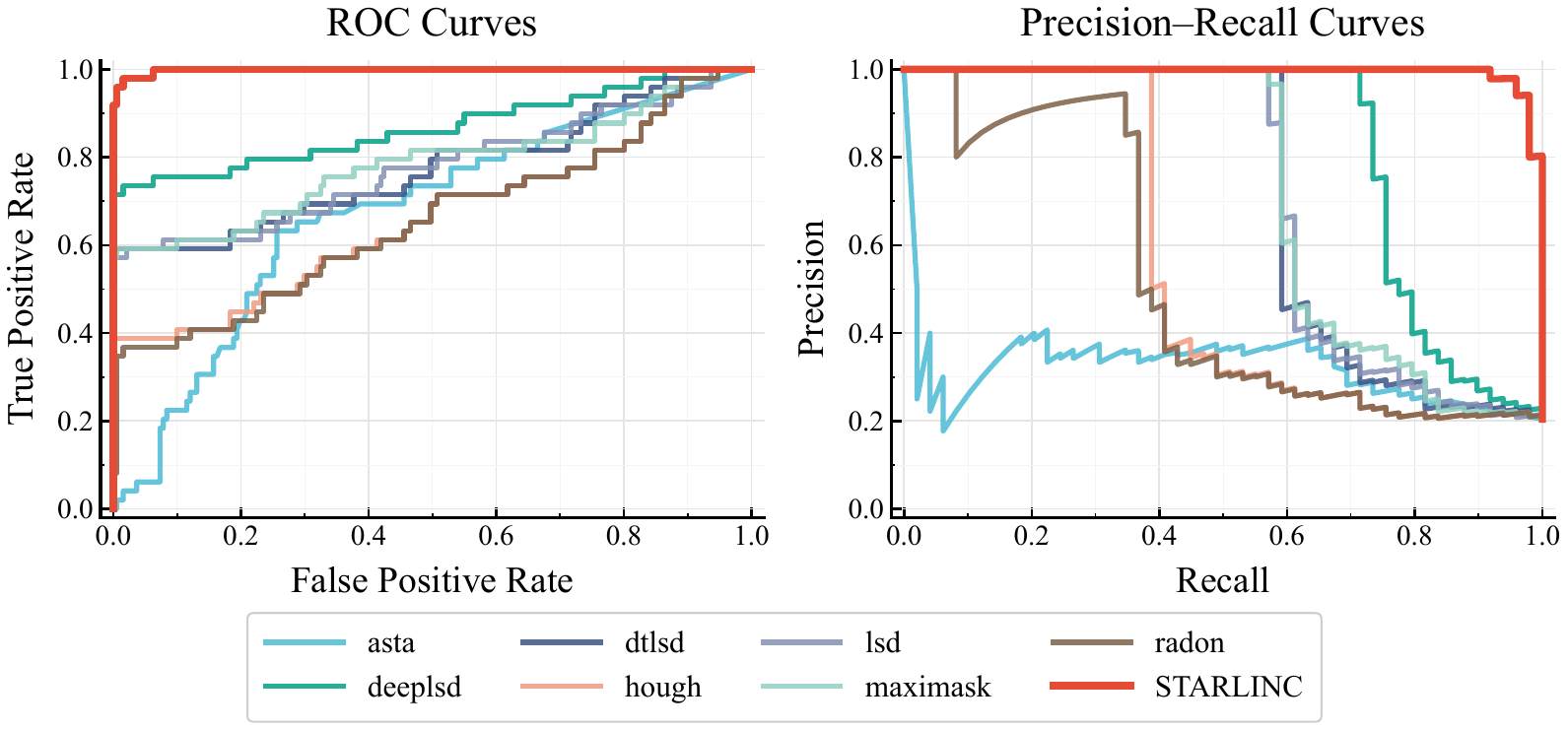}

\caption{Performance evaluation for classification model using (a) ROC curve and (b) PR curve showing performance across different decision thresholds.}

\label{fig:exp:roc_pr_cls}

\end{figure}

\begin{table}[]
\centering
\caption{Classification performance comparison. 
}
\label{tab:cls_perf}
\begin{tabular}{llcccccc}
\toprule
\multirow{2}{*}{Category} & \multirow{2}{*}{Method} & \multicolumn{6}{c}{Metrics} \\
\cmidrule(lr){3-8}
& & Acc & Pre & Recall & F1 & ROC-AUC & PR-AUC \\
\midrule
\multirow{3}{*}{Classical}

	&	Hough~\cite{hough}	&	0.875	&	0.913	&	0.429	&	0.583	&	0.692	&	0.756	\\
	&	Radon~\cite{radon}	&	0.863	&	0.944	&	0.347	&	0.507	&	0.671	&	0.708	\\
	&	LSD~\cite{lsd}	&	0.204	&	0.204	&	1.000	&	0.339	&	0.777	&	0.723	\\
	\midrule														
	\multirow{2}{*}{Learning-based}														
	&	DeepLSD~\cite{deeplsd}	&	0.204	&	0.204	&	1.000	&	0.339	&	0.868	&	0.834	\\
	&	DT-LSD~\cite{dt-lsd}	&	0.917	&	1.000	&	0.592	&	0.744	&	0.796	&	0.838	\\
	\midrule														
	\multirow{2}{*}{Trail detection}														
	&	ASTA~\cite{asta} 	&	0.429	&	0.244	&	0.857	&	0.380	&	0.677	&	0.384	\\
	&	MaxiMask~\cite{maximask}	&	0.913	&	0.967	&	0.592	&	0.734	&	0.795	&	0.830	\\
	\midrule
    \rowcolor{gray!15}
Proposed	&	\textbf{STARLINC (Ours)}	&	0.983	&	0.959	&	0.959	&	0.959	&	0.998	&	0.996	\\

\bottomrule
\end{tabular}
\end{table}

\subsection{Evaluation on Satellite Trail Classification}
\label{sec:eval_cls}

Beyond segmentation, we evaluate the classifier stage of \name, which serves as an upstream detector and provides image-level predictions and localization cues through a \gradcam to the segmentation module. 
\Cref{tab:cls_perf} summarizes the results, and Appendix (\Cref{fig:exp:roc_pr_cls}) provides the ROC and PR curves.

Overall, classical detectors exhibit a clear precision–recall trade-off. 
Hough and Radon achieve high precision (0.913/0.944) but low recall (0.429/0.347), while LSD over-triggers positives (precision 0.204). 
Learning-based baselines improve performance, with DT-LSD achieving the strongest results among them (F1 0.744), whereas DeepLSD shows over-detection consistent with the domain gap. 
Trail-specific methods are mixed: ASTA degrades on 7DT (F1 0.380) and MaxiMask is more competitive (F1 0.734) but remains limited in recall.

\name achieves the best performance across all metrics (accuracy 0.983, F1 0.959, ROC-AUC 0.998) with balanced precision and recall (0.959/0.959). 
We further provide ablation study of classification pipeline in \cref{tab:abl_cls_full2}.
The results show that removing synthetic trail generation and the inter-frame differential map degrades performance, while combining them yields the best results, indicating complementary benefits.

\subsection{Additional Experimental Results of the Classifier}
\label{app:sec:cls_sense}

\begin{table}[t]
\centering
\caption{Sensitivity analysis for classification performance.}
\label{tab:sensitivity_cls}
\renewcommand{\arraystretch}{1.2} %
\begin{tabular}{@{} l *{6}{c} @{}} %
\toprule
\multirow{2}{*}{\makecell[l]{Synthetic\\ratio (\%)}} & \multicolumn{6}{c}{Number of data} \\
\cmidrule(l){2-7} %
 & 150 & 300 & 450 & 600 & 750 & 900 \\
\midrule
30 & 0.651 & 0.863 & 0.835 & 0.863 & 0.882 & 0.865 \\
50 & 0.515 & 0.653 & 0.662 & 0.756 & 0.824 & 0.959 \\
70 & 0.626 & 0.626 & 0.492 & 0.364 & 0.492 & 0.545 \\
90 & 0.371 & 0.339 & 0.345 & 0.344 & 0.344 & 0.375 \\
\bottomrule
\end{tabular}
\end{table}

\begin{table}[t]
  \centering
  \caption{Ablation results of the classification task.}
  \label{tab:abl_cls_full2}
  \begin{tabular}{cccccc}
    \toprule
    \multicolumn{2}{c}{Method} & \multicolumn{4}{c}{Metrics} \\
    \cmidrule(lr){1-2}\cmidrule(lr){3-6}
    \S~\ref{sec:method:synth} & \S~\ref{sec:method:classification} & F1 & Acc & Pre & Recall \\
    \midrule
    \xmark & \xmark & 0.314 & 0.708 & 0.302 & 0.327 \\
    \cmark & \xmark & 0.381 & 0.783 & 0.457 & 0.327 \\
    \xmark & \cmark & 0.779 & 0.913 & 0.804 & 0.755 \\
    \cmark & \cmark & 0.959 & 0.983 & 0.959 & 0.959 \\
    \bottomrule
  \end{tabular}
\end{table}

\Cref{fig:exp:roc_pr_cls} further illustrates the classification performance using ROC and PR curves, complementing the quantitative results presented in \cref{sec:eval_cls}. 
The curves show that the proposed classifier maintains a favorable precision–recall trade-off across different decision thresholds.

\cref{tab:sensitivity_cls} presents the sensitivity analysis of the classification detector with respect to the synthetic data ratio and the number of training samples. 
All results are reported in terms of the F1 score. 
Overall, increasing the number of training samples generally improves performance, particularly when the dataset size increases from 150 to 450 samples, indicating that the classifier benefits from additional training data.

Among the tested configurations, a synthetic ratio of 50\% achieves the best overall performance, reaching the highest F1 score of 0.959 when 900 training samples are used. 
This suggests that a balanced mixture of real and synthetic samples provides the most effective training signal for the detector.

In contrast, higher synthetic ratios (e.g., 70\% and 90\%) lead to noticeably degraded performance across most dataset sizes. 
This indicates that excessive reliance on synthetic samples may introduce a distribution mismatch with real astronomical images, ultimately harming detection performance.
Based on these observations, we adopt a synthetic ratio of 50\% and 900 training samples as the default configuration for the classification detector.

\section{Synthetic Fraction Sensitivity}
\label{app:sec:synthetic_fraction}

\cref{tab:synthetic_fraction} reports the sensitivity analysis of the segmentation model with respect to the fraction of synthetic trail images among total trail images.
This quantity mainly controls the class balance between trail and non-trail images.
To directly evaluate the effect of synthetic trails, we vary the synthetic fraction among trail-containing images and report the segmentation performance in Table~\ref{tab:synthetic_fraction}. 
The reference \name setting corresponds to a synthetic fraction of 57.3\%.

\begin{table}[t]
\centering
\caption{
Sensitivity to the synthetic fraction among trail-containing training images.
}
\label{tab:synthetic_fraction}

{
\renewcommand{\arraystretch}{1.18}
\setlength{\tabcolsep}{4pt}
\resizebox{\linewidth}{!}{
\begin{tabular}{lcccccccc}
\toprule
Synthetic fraction (\%)
& 57.3
& 65.1
& 70.0
& 75.1
& 80.0
& 85.1
& 90.0
& 95.1 \\
\midrule
mIoU 
& \makecell[c]{$0.485$\\$\pm0.012$}
& \makecell[c]{$\mathbf{0.521}$\\$\boldsymbol{\pm}\mathbf{0.018}$}
& \makecell[c]{$0.493$\\$\pm0.014$}
& \makecell[c]{$0.492$\\$\pm0.008$}
& \makecell[c]{$0.504$\\$\pm0.010$}
& \makecell[c]{$0.506$\\$\pm0.017$}
& \makecell[c]{$0.472$\\$\pm0.012$}
& \makecell[c]{$0.504$\\$\pm0.008$} \\
\bottomrule
\end{tabular}
}
}

\end{table}

The results show that \name is not highly sensitive to the fraction of synthetic trails among trail-containing images. 
Even at 95.1\% synthetic trails, the model achieves 0.504 mIoU, which is comparable to the reference setting. 
These results support the use of synthetic trail injection as a scalable source of pixel-level supervision. 
At the same time, the synthetic trail realism study in \S~\ref{sec:additional_baselines} shows that improving the rendering model, such as by adding PSF-aware trail profiles, can further improve real-image segmentation performance.

\section{Fine-tuned Baseline Results}
\label{sec:finetuned_baselines}

For completeness, we report the 7DT results of the fine-tuned learning-based baselines.%
The baselines include DeepLSD, DT-LSD, ASTA, and MaxiMask, which are adapted to the target 7DT setting.
As shown in \cref{tab:finetuned_baselines}, fine-tuning improves the applicability of existing line or trail detectors to astronomical images, but their performance remains below STARLINC.
These results indicate that adapting existing detectors alone is insufficient for robust pixel-level satellite-trail segmentation on 7DT observations.

\begin{table}[t]
\centering
\caption{Results of fine-tuned learning-based baselines on the 7DT dataset.}
\label{tab:finetuned_baselines}
\resizebox{0.5\linewidth}{!}{%
\begin{tabular}{lcccc}
\toprule
Method & mIoU & Dice & Precision & Recall \\
\midrule
DeepLSD$^\dagger$  & 0.269 & 0.380 & 0.293 & 0.628 \\
DT-LSD$^\dagger$   & 0.381 & 0.474 & 0.482  & 0.477 \\
ASTA$^\dagger$     & 0.285 & 0.394 & 0.360 & 0.506 \\
MaxiMask$^\dagger$ & 0.220 & 0.289 & 0.734 & 0.180 \\
\bottomrule
\end{tabular}}
\end{table}

\section{Additional Motivational Results on Low-SNR Images}
\label{app:snr}

In this section, we present qualitative examples of how decreasing SNR degrades line detection.
To generate images with controlled SNR levels, we inject synthetic noise into the original images. 
Specifically, background noise is modeled using a Gaussian component, and the noise magnitude is adjusted to achieve a desired target SNR level. 
For each image, a target SNR ($\mathrm{SNR}_{\text{target}}$) is specified, and the noise standard deviation ($\sigma_n$) is tuned such that the effective SNR of the resulting image approximates $\mathrm{SNR}_{\text{target}}$.
This procedure allows us to systematically construct low-SNR samples that mimic challenging observational conditions frequently encountered in astronomical imaging, such as weak linear structures and increased background fluctuations.

\cref{fig:snr_example} shows predictions of DeepLSD~\cite{deeplsd} on the Wireframe dataset~\cite{huang2018learning} under different SNR levels, 
alongside ground truth annotations.
As the SNR decreases, structural line patterns become increasingly difficult to distinguish from background noise, causing the detector to miss valid segments and produce fragmented or unstable predictions. 
These results highlight the inherent difficulty of reliably detecting lines in low-SNR environments. 

While modern line detectors perform well on natural images with relatively clean edges, their accuracy degrades substantially when the signal is weak relative to the background noise, a condition frequently encountered in astronomical observations. 
This motivates the development of methods robust to low-SNR conditions for satellite trail detection.

\section{Network Size Sensitivity}
\label{app:sec:unet_size}

We evaluate the sensitivity of \name to the size of the U-Net segmentation backbone by varying the number of parameters while keeping the rest of the pipeline fixed.

\begin{table}[t]

\centering

\caption{U-Net size sensitivity analysis.}

\label{tab:unet_size_sensitivity}

\resizebox{0.9\columnwidth}{!}{%

\setlength{\tabcolsep}{2pt}

\begin{tabular}{ccccccccc}

\toprule

\# Params(M)
& 3.67
& 4.67
& 5.94
& 6.82
& \cellcolor{blue!8}\textbf{7.76 (Ours)}
& 8.96
& 10.97
& 20.61 \\

\midrule

mIoU
& 0.475
& 0.480
& 0.480
& 0.488
& \cellcolor{blue!8}0.485
& 0.480
& 0.483
& 0.473 \\

\bottomrule

\end{tabular}

}

\end{table}

As shown in Table~\ref{tab:unet_size_sensitivity}, the mIoU remains stable across the tested U-Net sizes, varying only from 0.473 to 0.488, with only minor drops at the smallest and largest scales.
This stability suggests that the performance is driven more by the three-channel input representation than by increasing the segmentation backbone size, and we therefore use the 7.76M-parameter U-Net as the default setting.

\section{Effect of Input Bit Depth}
\label{app:sec:bit_depth}

We further examine whether the 8-bit preprocessing used in the main pipeline limits the detection of faint satellite trails. 
This concern is important because astronomical CCD images are commonly stored with higher bit depth, whereas display-oriented 8-bit conversions may compress or clip low-contrast flux variations. 
Such loss could be especially harmful for trails close to the noise
floor.

We focus on detectable faint trails with SNR \(>3\), following the common observational convention that signals below this level are close to the sensitivity limit of ground-based telescopes and are difficult to distinguish reliably from background noise~\cite{7dt}.
This threshold defines the regime where a visual or algorithmic trail label is meaningful, rather than a limitation specific to \name.

To test whether bit-depth conversion is the dominant performance bottleneck in our setting, we evaluate a 16-bit variant of \name
using the same overall training and inference pipeline. 
As shown in Table~\ref{tab:bit_depth}, the 16-bit variant achieves comparable but slightly lower performance than the main 8-bit setting. 
This suggests that bit depth alone does not explain the remaining missed trail pixels in our current pipeline. 
One possible reason is that long, coherent linear structures remain detectable after 8-bit scaling when their SNR is above the target threshold, whereas other factors, such as residual stellar structures, seeing variation, and the training distribution, also affect segmentation quality.

We therefore use the 8-bit setting as the reference configuration for the main experiments, while treating FITS-native or fully 16-bit processing as an important direction for future work. 
In particular, future pipelines may benefit from jointly optimizing bit-depth handling, normalization, and faint-trail supervision rather than changing the input representation alone.
The predicted masks from the main 8-bit pipeline can also be applied directly to the original FITS images, so the 8-bit representation is used for mask prediction rather than for replacing the scientific data product.

\begin{table}[t]
\centering
\caption{
Effect of input bit depth.
}
\label{tab:bit_depth}
\resizebox{0.70\linewidth}{!}{
\begin{tabular}{lcccc}
\toprule
Setting & mIoU  & Dice  & Precision & Recall \\
\midrule
\name{}-8bit(ours) & \textbf{0.485} & \textbf{0.653} & 0.815 & \textbf{0.545} \\
\name{}-16bit & 0.450 & 0.621 & \textbf{0.816} & 0.501 \\
\bottomrule
\end{tabular}
}
\end{table}

\section{Data Preservation through Pixel-Level Trail Removal}

\begin{minipage}[]{0.66\textwidth}
Conventional survey pipelines often discard entire exposures once satellite trails are detected. 
Although this prevents contamination from propagating into downstream analysis, it also leads to substantial data loss because valid astronomical information outside the trail region is removed together with the artifact. 
\end{minipage}
\hfill
\begin{minipage}[]{0.3\textwidth}
\vspace{-20mm}
\centering
\captionof{table}{Comparison of data preservation ratio}
\begin{tabular}{lcc}
\toprule
 & Baseline & Ours \\
\midrule
Ratio & 79.58\% & 99.95\% \\
\bottomrule
\end{tabular}
\label{tab:data_preserve}
\vspace{-15mm}
\end{minipage}
As discussed in \cref{sec:intro}, the number of LEO satellites is rapidly increasing, implying that the fraction of contaminated exposures will continue to grow. 
Under such conditions, image-level removal becomes increasingly inefficient, as a larger portion of the observational data would be discarded.
In contrast, the proposed method removes satellite trails at the pixel level, allowing uncontaminated regions of the image to remain usable for scientific analysis. 
To quantify this effect, we measure the preservation ratio defined as the fraction of pixels that remain usable after artifact removal. 
\cref{tab:data_preserve} compares this metric between conventional image removal and the proposed approach. 
While the conventional strategy retains only about 79.58\% of the data due to full-image rejection, our method preserves approximately 99.95\% of valid pixels by selectively masking the trail region. 
This difference highlights the practical advantage of pixel-level artifact removal, which becomes increasingly important for future large-scale surveys where the frequency of satellite contamination is expected to rise.

\section{Limitations}

\label{app:sec:limitation}

In this work, we evaluate \name using the available astronomical datasets containing satellite trails. 
However, the evaluation scope is currently limited by the availability of public datasets. 
Many modern astronomical observations remain proprietary to the survey teams that collected them and are typically released only after a proprietary period.
As a result, comprehensive evaluation across a wider range of observational datasets is still limited at present. 
As more astronomical survey become publicly available in the future, broader validation across diverse astronomical observations would further demonstrate the applicability of \name.

\begin{figure}[t]
\centering
\includegraphics[width=\linewidth]{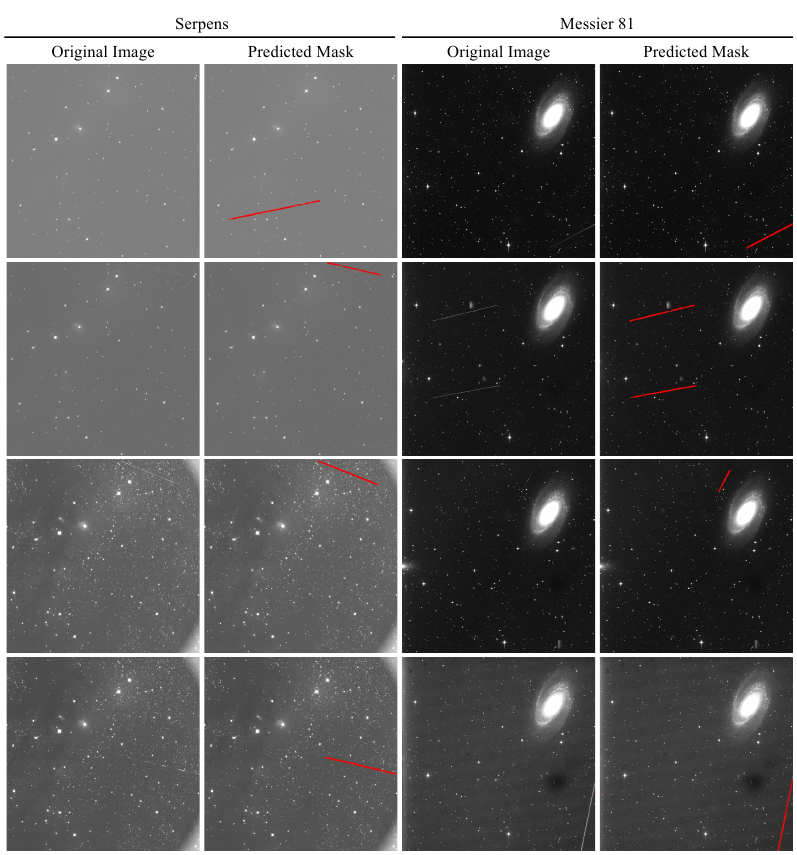}
\caption{Applying \name to other astronomical datasets (Serpens and Messier 81). The predicted masks accurately localize satellite trails while preserving surrounding celestial structures.}
\label{fig:other_data}
\end{figure}

\begin{figure}[p]
\centering
\includegraphics[width=\linewidth]{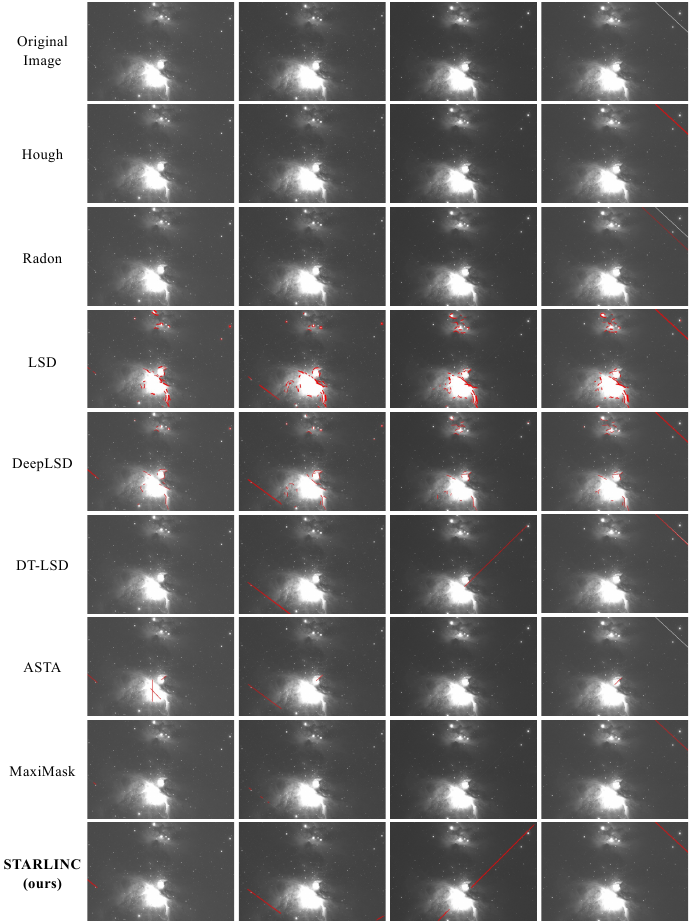}
\caption{Additional qualitative comparison of satellite trail detection and removal results across different methods.}
\label{fig:results_full1}
\end{figure}

\begin{figure}[p]
\ContinuedFloat
\centering
\includegraphics[width=\linewidth]{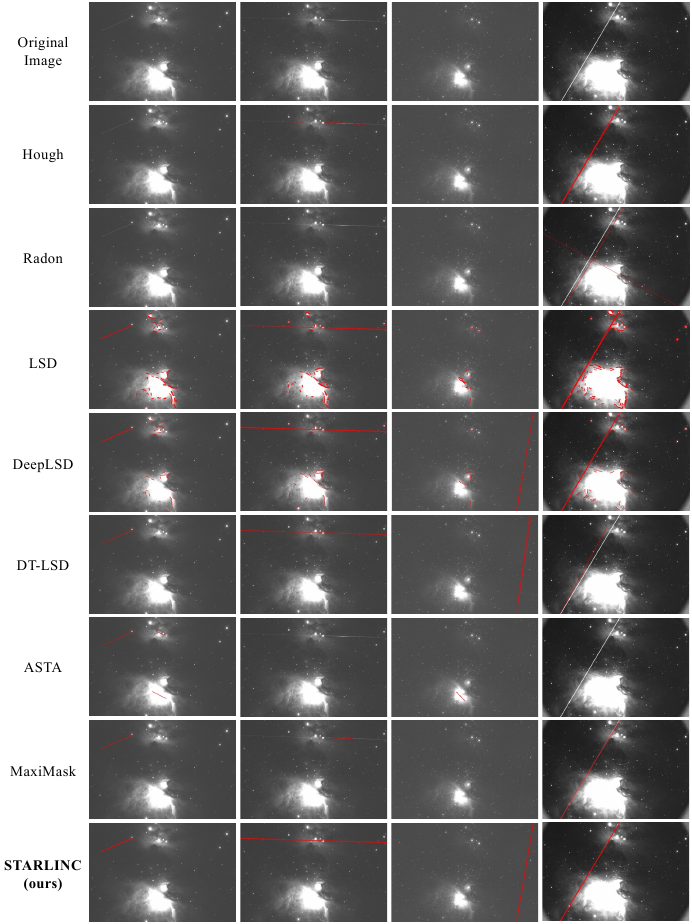}
\caption{Additional qualitative comparison of satellite trail detection and removal results across different methods (continued).}
\label{fig:results_full2}
\end{figure}

\begin{figure}[h]
  \centering
  \includegraphics[width=\textwidth]{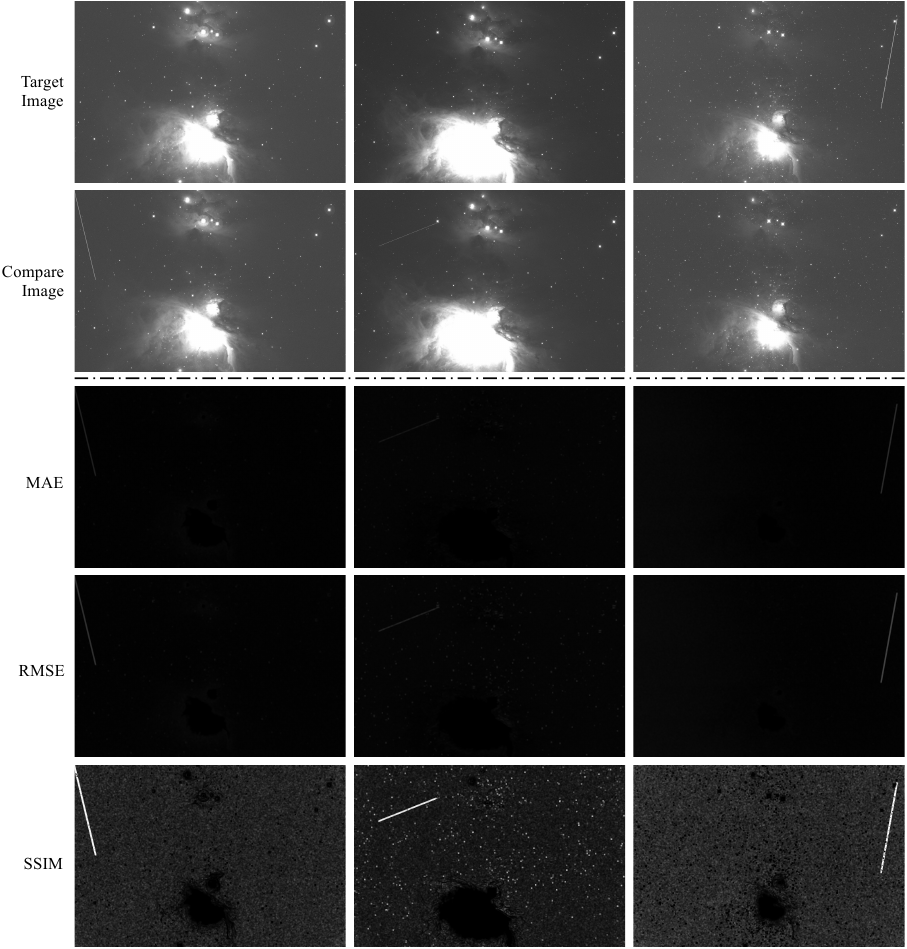}
  \caption{Comparison of different distance metrics for inter-frame differential map computation. Given a target image and another comparison image from the same observation group, we compute difference maps using MAE, RMSE, SSIM.}
  \label{fig:diff}
\end{figure}

\begin{figure}[p]
    \centering
    \includegraphics[width=\linewidth]{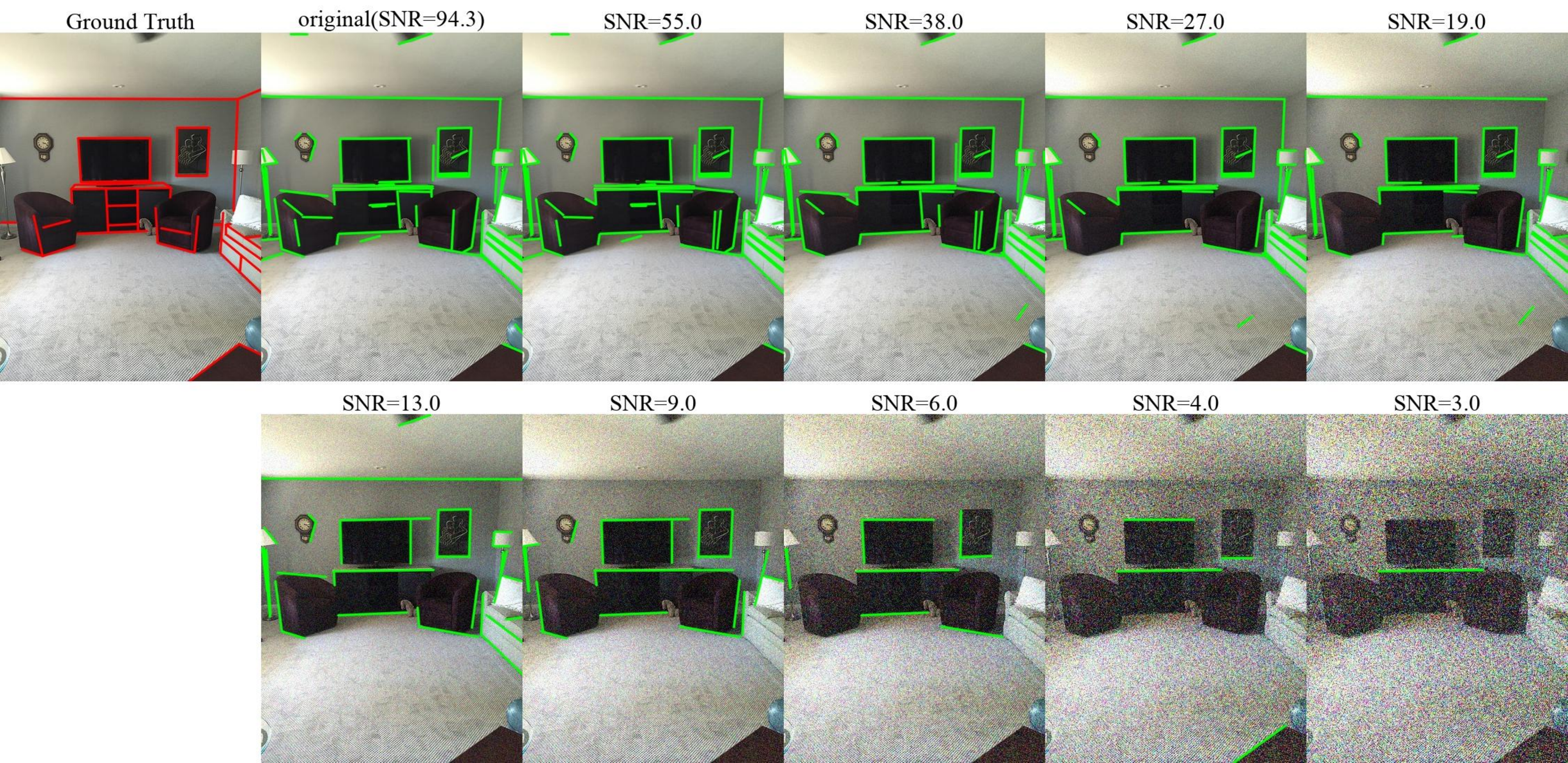}
    \vspace{2mm}
    \includegraphics[width=\linewidth]{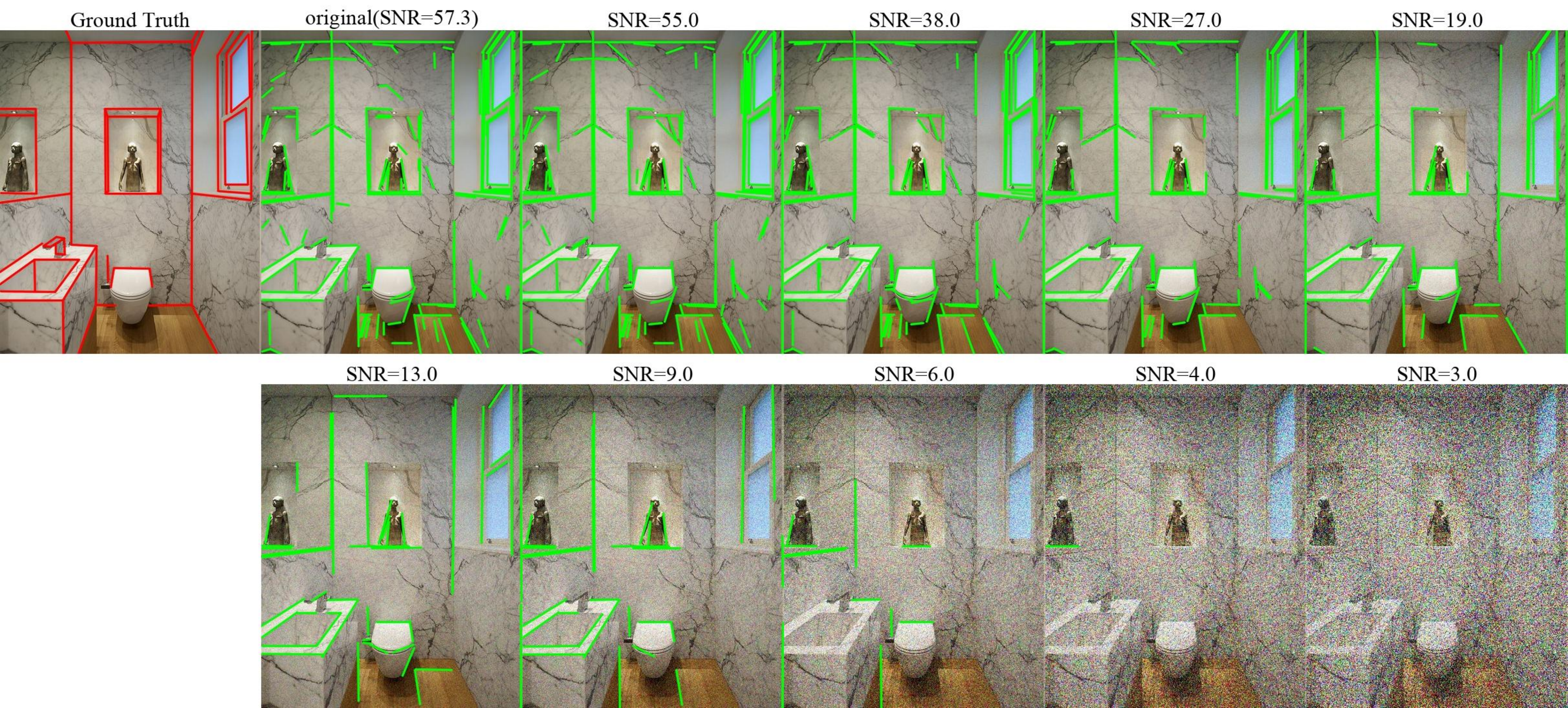}
    \vspace{2mm}
    \includegraphics[width=\linewidth]{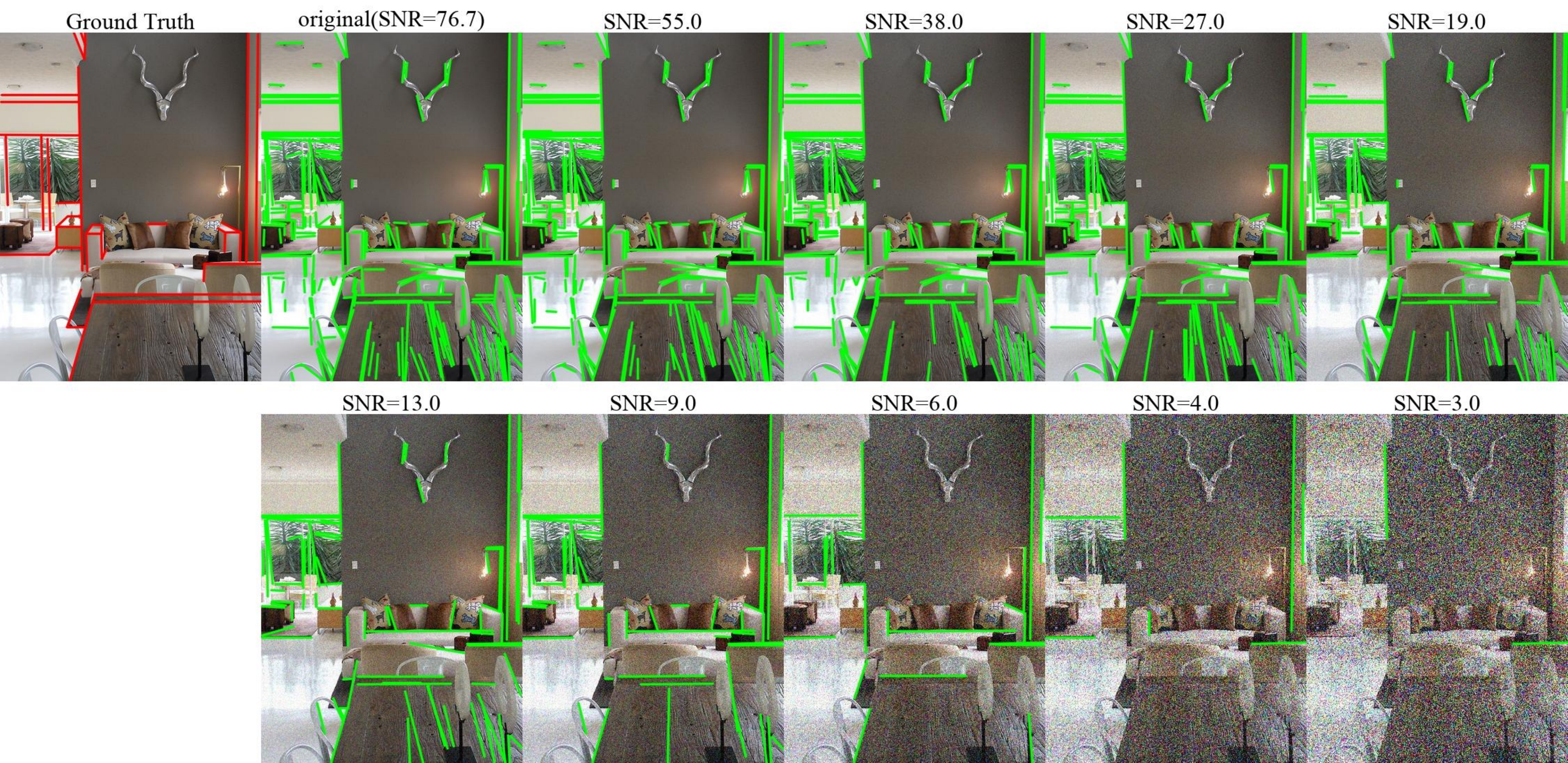}
    \caption{Example predictions of DeepLSD~\cite{deeplsd} on the Wireframe dataset~\cite{huang2018learning} under different SNR levels, compared with ground-truth.}
    \label{fig:snr_example} 
\end{figure}

\end{document}